\documentclass[11pt, a4paper, logo, onecolumn]{berkeley}

\usepackage[all]{hypcap}
\usepackage{xspace}
\usepackage[capitalize,noabbrev]{cleveref}
\usepackage{subcaption}
\usepackage{wrapfig}
\usepackage{lipsum}
\usepackage{multirow}
\usepackage{tabularx}

\usepackage{url}
\usepackage{float}
\usepackage{tikz}
\usepackage{tabularx} % Add this to your preamble if not already included
\usepackage{array}
\newcolumntype{Y}{>{\centering\arraybackslash}X}
\usepackage{multirow} 
\usepackage{makecell}
\usepackage{subcaption}
\usepackage{hyperref}
\usepackage{amsmath}

\makeatletter
\def\mathcolor#1#{\@mathcolor{#1}}
\def\@mathcolor#1#2#3{%
  \protect\leavevmode
  \begingroup
    \color#1{#2}#3%
  \endgroup
}
\makeatother

\title{Precise and Dexterous Robotic Manipulation via
Human-in-the-Loop Reinforcement Learning}
\runningtitle{HIL-SERL: Precise and Dexterous Robotic Manipulation via
Human-in-the-Loop Reinforcement Learning}
\reportnumber{} % Leave blank if n/a 

\renewcommand{\today}{Oct 2024} 

\author[1]{Jianlan Luo}
\author[1]{Charles Xu}
\author[1]{Jeffrey Wu}
\author[1]{Sergey Levine}

\affil[1]{Department of Electrical Engineering and Computer Sciences, UC Berkeley}

\correspondingauthor{Jianlan Luo(\href{mailto:jianlanluo@berkeley.edu}{jianlanluo@berkeley.edu}), Charles Xu(\href{mailto:xuc@berkeley.edu}{xuc@berkeley.edu}).}

\begin{abstract}
	Reinforcement learning (RL) holds great promise for enabling autonomous acquisition of complex robotic manipulation skills, but realizing this potential in real-world settings has been challenging. We present a human-in-the-loop vision-based RL system that demonstrates impressive performance on a diverse set of dexterous manipulation tasks, including dynamic manipulation, precision assembly, and dual-arm coordination. Our approach integrates demonstrations and human corrections, efficient RL algorithms, and other system-level design choices to learn policies that achieve near-perfect success rates and fast cycle times within just 1 to 2.5 hours of training. We show that our method significantly outperforms imitation learning baselines and prior RL approaches, with an average 2x improvement in success rate and 1.8x faster execution. Through extensive experiments and analysis, we provide insights into the effectiveness of our approach, demonstrating how it learns robust, adaptive policies for both reactive and predictive control strategies. Our results suggest that RL can indeed learn a wide range of complex vision-based manipulation policies directly in the real world within practical training times. We hope this work will inspire a new generation of learned robotic manipulation techniques, benefiting both industrial applications and research advancements. Videos and code are available at our project website \url{https://hil-serl.github.io/}.
\end{abstract}

\begin{document}

\maketitle

\section{Introduction}\label{sec:intro}

Manipulation is one of the foundational problems in robotics, and achieving human-level performance on dynamic, dexterous manipulation tasks is a longstanding pursuit in the field~\cite{Cui2021TowardNL}.
Reinforcement learning (RL) holds the promise of enabling autonomous acquisition of complex and dexterous robotic skills. By learning through trial and error, an effective RL method should in principle be able to acquire highly proficient skills that are tailored to the particular physical characteristics of the deployment task. This could result in performance that not only exceeds that of hand-designed controllers but also surpasses human teleoperation. However, realizing this promise in real-world settings has been challenging due to issues with sample complexity, assumptions (e.g., accurate reward functions), and optimization stability. 
RL methods have been effective for training in simulation~\cite{hwangbo2019sr, hwangbo2020sr, chen2024visualdexterity, loquercio2021drone}, and for training on existing large real-world datasets with the aim of broad generalization~\cite{kalashnikov2018qtopt, kalashnikov2021mtopt}. They have also been used with hand-designed features or representations for narrowly tailored tasks~\cite{theodorou2010generalized, chebotar2016pigps}. However, developing general-purpose vision-based methods that can efficiently acquire physically complex skills, with proficiency exceeding imitation learning and hand-designed controllers, has been comparatively difficult. 
We believe that making fundamental progress on this front can unlock new opportunities, which will then enable the development of truly performant robotic manipulation policies.

In this paper, we develop a reinforcement learning (RL) system for vision-based manipulation that can acquire a wide range of precise and dexterous robotic skills. 
Our system, named Human-in-the-Loop Sample-Efficient Robotic Reinforcement Learning (HIL-SERL), addresses the previously mentioned challenges by integrating a number of components that enable fast and highly performant vision-based RL in the real world. 

\begin{figure}[t!]
    \centering
    \includegraphics[width=0.9\textwidth]{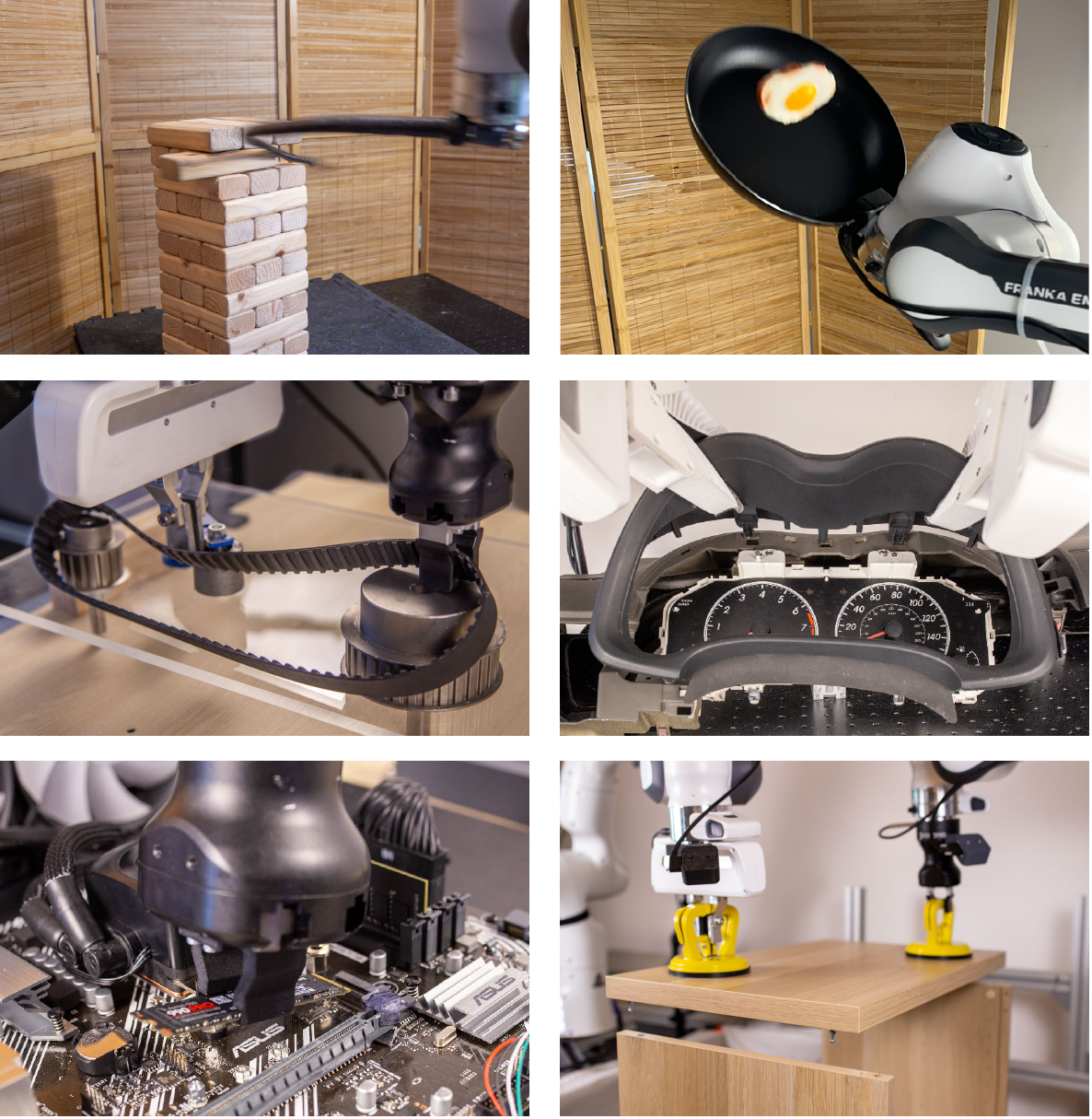}
    \caption{\textbf{Overview of experimental tasks.} A subset of tasks considered in this paper, they include whipping out a Jenga block from its tower, flipping an object in a pan, assembling complex devices such as a timing belt, a dashboard, a motherboard, and an IKEA shelf.}
    \label{fig:teaser}
\end{figure}
To address the optimization stability issue, we employ a pretrained visual backbone for policy learning. To handle the sample complexity issue, we utilize a sample-efficient off-policy RL algorithm based on RLPD~\cite{ball2023efficient} that also incorporates human demonstrations and corrections. Additionally, a well-designed low-level controller is included to ensure safety during policy training.
During training, the system queries a human operator for potential corrections, which are then used to update the policy in an off-policy manner. We found this human-in-the-loop correction procedure is crucial for enabling the policy to learn from its mistakes and improve performance, particularly for challenging tasks considered in this paper, which are hard to learn from scratch.

As shown in Fig.~\ref{fig:teaser}, the tasks our system solves include dynamically flipping an object in a pan, whipping out a Jenga block from a tower, handing over objects between two arms, and assembling complex devices such as a computer motherboard, an IKEA shelf, a car dashboard, or a timing belt, using either one or two robotic arms.
These tasks present significant challenges in terms of complex and intricate dynamics, high-dimensional state and action spaces, long horizons, or combinations thereof. 
Some of these skills were previously considered infeasible to train with RL directly in real-world settings, such as many of the dual-arm manipulation tasks, or nearly insurmountable with current robotics methods, like timing belt assembly or Jenga whipping. And they require different types of control strategies, such as reactive closed-loop control for precise manipulation tasks or delicate open-loop behaviors that are otherwise very difficult to prescribe, e.g., Jenga whipping.
However, perhaps the most surprising finding is that our system can train RL policies to achieve near-perfect success rates and super-human cycle times on almost all of the tasks with only one hour to 2.5 hours of training time in the real world. Our trained RL policies greatly outperform imitation learning methods that are trained on the same amount of human data, e.g., same number of episodes of demonstrations or corrections, on average 101\% improvement in terms of success rate and 1.8x faster in cycle time. 
The result is significant because it demonstrates that RL can indeed learn a wide range of complex vision-based manipulation policies directly in the real world within practical training times, which was previously considered infeasible with earlier methods. Moreover, RL does so with a superhuman level of performance that greatly exceeds that of imitation learning and hand-designed controllers.

To assess the effectiveness of our system, we compare it against several state-of-the-art RL methods and conduct ablation studies to understand the contribution of each component. The results demonstrate that our system not only outperforms the relevant baselines but also highlights that the impressive empirical results are indeed due to the careful integration of these components. 
Additionally, we provide a comprehensive analysis of the empirical results, offering insights into the effectiveness of RL-based manipulation. This analysis explores why RL achieves a near-perfect success rate, and further examines the flexibility of RL policies to serve as a general-purpose vision-based policy for acquiring distinct types of control strategies. 

In summary, our contributions demonstrate that with the appropriate system-level design choices, RL can effectively solve a wide range of dexterous and complex vision-based manipulation tasks in the real world. Notably, our system is, to the best of our knowledge, the first to achieve dual-arm coordination with image inputs using RL in real-world settings, as well as tasks like whipping out a Jenga block and assembling a timing belt. We also provide a comprehensive analysis of the empirical success of RL-based manipulation, offering insights into the effectiveness of RL-based manipulation. This analysis shapes our understanding of why RL succeeds in these complex tasks and suggests potential directions for further extending RL-based manipulation to even more challenging scenarios. 

With the results presented in this paper, we hope this work will serve as a stepping stone for future learning-based robotic manipulation research, and in the long term, could enable robust deployable robotic manipulation skills capable of adapting to diverse environments and tasks, bringing us closer to the goal of general-purpose robotic manipulation.

\section{Related Work}\label{sec:related}
The proposed system uses RL to solve dexterous manipulation tasks, thus we survey related works on real-world robotic RL methods and systems, as well as other approaches that address similar dexterous manipulation tasks. 

\paragraph{Algorithms and systems for real-world RL} 
Real-world robotic reinforcement learning (RL) requires algorithms that are sample-efficient in handling high-dimensional inputs such as onboard perception and supporting easy specification of rewards and resets. 
Several algorithms have demonstrated the ability to learn efficiently directly in the real world~\citep{riedmiller2009reinforcement,levine16gps,Luo-RSS-21, yang2020data, zhan2021framework, tebbe2021sample, popov2017data, luo2019reinforcement, zhao2022insertion, hu23reboot, residualrl,hu2024ibrl, Rajeswaran-RSS-18, visual_residual_rl, luo2024serl}. These include variants of off-policy RL~\citep{kostrikov23wip, hu23reboot, luo2023rlif}, model-based RL~\citep{hester2013texplore, wu22daydreamer, nagabandi19pddm, rafailov21orllatent, luo2018deep}, and on-policy RL~\citep{zhu19dexterous}. Despite progress, these often require long training times. Our system achieves super-human performance on complex tasks with shorter training times.
Other works have been researching on inferring rewards from raw visual observation through success classifiers ~\citep{fu2018variational, li21mural}, foundation-model-based rewards ~\citep{du2023vision, mahmoudieh22zeroshot, fan22minedojo}, and rewards from videos~\citep{ma23vip, ma23liv}. 
Additionally, to enable autonomous training, there have been a number of algorithmic advances in reset-free learning~\citep{gupta21mtrf, sharma2021Autonomous, zhu20r3l, xie22help, sharma23sir} that require minimal human interventions during training. While we do not introduce new algorithms in these areas, our framework effectively integrates existing methods. As detailed in the methods section, using binary classifier-based rewards with demonstrations is effective for the complex tasks in this paper.

One of the most relevant works to our research is SERL~\citep{luo2024serl}, which also presents a system for training reinforcement learning (RL) policies for manipulation tasks. Our approach differs from SERL in that: we incorporate both human demonstrations and corrections to train RL policies, whereas SERL relies solely on human demonstrations. While this might appear to be a minor distinction, our results indicate that integrating corrections is crucial for enabling the policy to learn from its mistakes and improve performance, especially for tasks that are challenging for the agent to learn from scratch. Additionally, SERL focuses on simpler tasks with relatively short horizons and does not address dual-arm coordination or dynamic manipulation. Our unique contribution is demonstrating that our approach can effectively learn general-purpose vision-based manipulation policies across a wide range of tasks with varying physical characteristics, setting our system fundamentally different from prior work on SERL.

\paragraph{Dexterous robotic manipulation} 
For some of the tasks considered in this paper, prior works have explored alternative approaches. 
In insertion tasks, prior works have used model-based approaches~\citep{tang2016autonomous, jin2021contact} and end-effector tooling mechanisms with passive compliance~\citep{morgan2021compliant, su2022insertion}. These methods often rely on state-based models without perception or require task-specific development, limiting robustness and adaptability.
Another approach involves using visual servoing in a multi-stage pipeline to align the robotic arm with the target, followed by search primitives for insertions~\citep{spector2022insertionnet, Chang2024insertone,usbvsinsertion}. They also face challenges with feature reliability and alignment precision.
In contrast, our method employs a much tighter perception-action loop. It learns task-relevant visual features and visuomotor policies in a closed-loop manner, crucial for many of the reactive high-precision tasks. The learned policy can be viewed as an instance of output feedback control from the controls perspective~\citep{murray2008feedback}. 

There are also a number of works on tackling the dynamic manipulation tasks~\citep{mason1993dynamic} considered in this paper. 
\citet{kormushev2010pancake} utilized a motion capture system and dynamic motion primitives~\citep{dmp} to learn flipping an object in the pan. However, our system directly uses pixel inputs, which alleviates the need for precise motion capture systems while achieving significantly higher success rates.    
\citet{fazeli2019jenga} proposed a learning method to push out Jenga block from its tower in a quasi-dynamic manner. Our approach, however, employs a whip to dynamically remove the Jenga block, presenting a more challenging task that requires a much more sophisticated control policy. 
Additionally, while there are studies on flexible object manipulation, such as cable routing~\citep{luo2024cable, jin2019cable}, tracing, or untangling~\citep{viswanath2023cabletracing,Shivakumar2023cable, viswanath2022autonomous}, the timing belt assembly task in our paper demands reactive yet precise coordination between two arms to dynamically adjust both the tensioner and the timing belt. This task is fundamentally different and more challenging than previous works on cable manipulation.
\section{Human-in-the-Loop Reinforcement Learning System}\label{sec:method}

In this section, we provide a detailed description of the methods used in the paper. For an animated movie summarizing the presented methods, please refer to the accommodating video.

\subsection{Preliminaries and Problem Statement}
Robotic reinforcement learning tasks can be defined via an MDP $\mdp = \{\states, \actions, \rho, \transition, \reward, \gamma\}$, where $\bs \in \states$ is the state observation (e.g., an image in combination with the robot's proprioceptive state information), $\ba \in \actions$ is the action (e.g., the desired end-effector twist), $\rho(\bs_0)$ is a distribution over initial states, $\transition$ is the unknown and potentially stochastic transition probabilities that depend on the system dynamics, and \mbox{$\reward: \states \times \actions \rightarrow \mathbb{R}$} is the reward function, which encodes the task. An optimal policy $\pi$ is one that maximizes the cumulative expected value of the reward, i.e., $E[\sum_{t=0}^H \gamma^t r(\bs_t,\ba_t)]$, where the expectation is taken with respect to the initial state distribution, transition probabilities, and policy $\pi$. In practice, the policy $\pi(\ba|\bs)$ is usually modeled as a Gaussian distribution parameterized by a neural network.

To implement reinforcement learning algorithms for robotic tasks, we must carefully select appropriate state observation spaces $\states$ and action spaces $\actions$. This involves choosing the right combination of cameras, proprioceptive states, and corresponding robot low-level controllers.
For all our tasks, we employ a sparse reward function. This function makes a binary decision on whether a task is successful or not using a trained classifier. In this setup, the optimization objective $E[\sum_{t=0}^H \gamma^t r(\bs_t,\ba_t)]$ aims to maximize the probability of success for each trajectory. Ideally, at convergence, the policy should succeed at every attempt.

% To implement reinforcement learning algorithms in real robotic systems, the MDP action $\ba$ must interface with the robot's low-level controller, typically presented as setpoints. 

Specifically, the core underlying RL algorithm that we build on is RLPD~\citep{ball2023efficient}, which we chose for its sample efficiency and ability to incorporate prior data. At each training step, RLPD samples equally between prior data and on-policy data to form a training batch~\citep{song2023hybrid}. It then updates the parameters of a parametric Q-function $Q_\phi(\bs,\ba)$ and the policy $\pi_\theta(\ba|\bs)$ according to the gradient of their respective loss functions:

\begin{align}
    \mathcal{L}_Q(\phi) &\!=\! E_{\bs,\ba,\bs'}\!\!\left[ \!\left( Q_{\phi}(\bs,\ba) \!-\! \left( r(\bs,\ba) \!+\! \gamma E_{\ba'\sim \pi_{\theta}}[Q_{\bar{\phi}}(\bs',\ba')] \right) \right)^2\! \right]\\
    \mathcal{L}_\pi(\theta) &\!=\! -E_{\bs}\left[ E_{\ba \sim \pi_\theta(\ba)}[Q_\phi(\bs,\ba)] + \alpha \mathcal{H}(\pi_\theta(\cdot | \bs) \right],
    \end{align}
    \noindent where $Q_{\bar{\phi}}$ is a \emph{target network}~\citep{mnih2013playing}, and the actor loss uses entropy regularization with an adaptively adjusted weight $\alpha$~\citep{haarnoja2018soft}.

\subsection{System Overview}
\begin{figure}[t!]
    \centering
    \includegraphics[width=\textwidth]{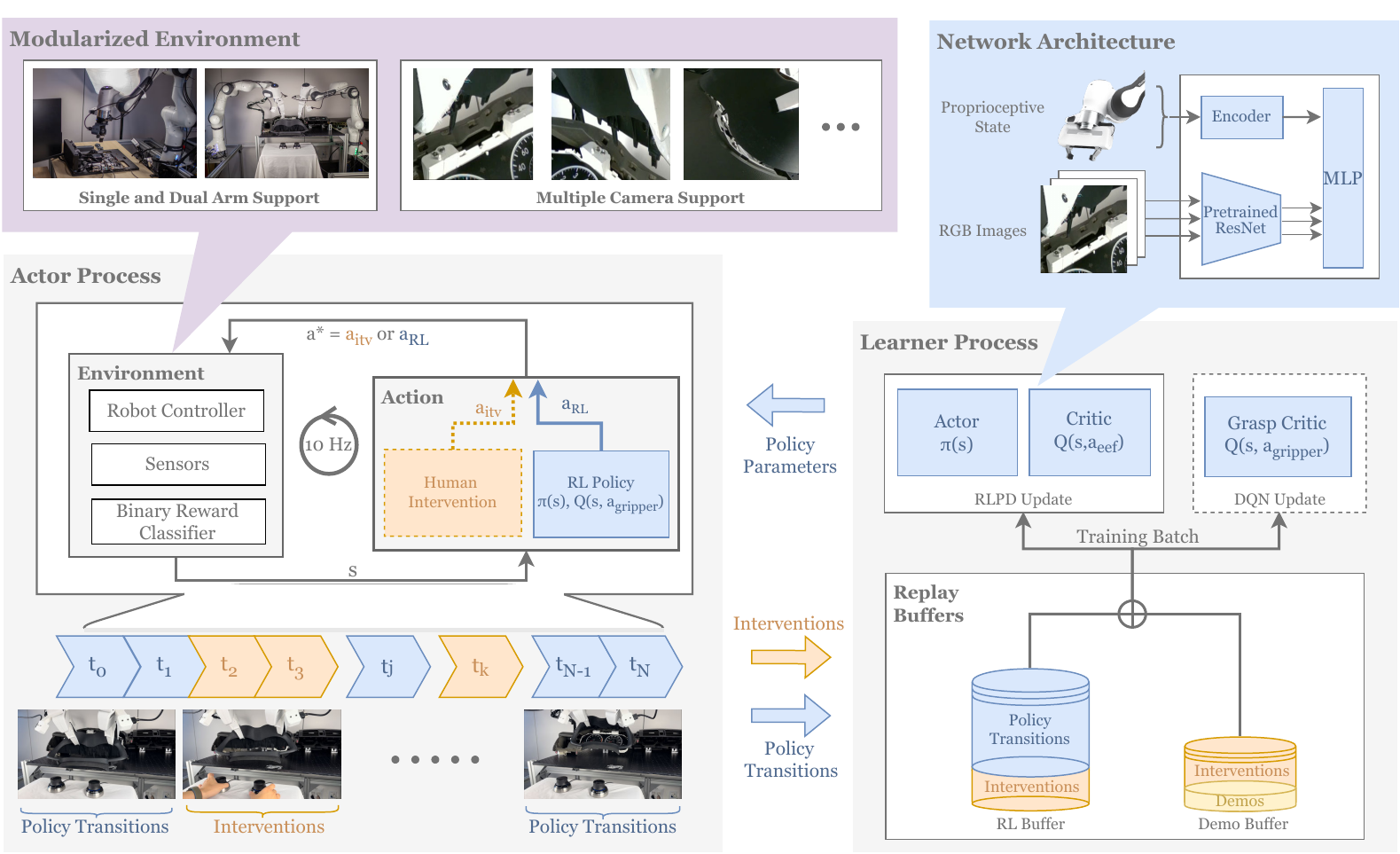}
    \caption{\textbf{Overview of HIL-SERL.} 
This figure illustrates the architecture of HIL-SERL, which comprises three primary components: the actor process, the learner process, and replay buffers. These components communicate asynchronously to facilitate efficient data flow. The actor process receives updated policy parameters from the learner process, interacts with the environment, and sends collected interaction data to the replay buffers. The environment is modular, supporting various external devices and multiple robotic arms. A human operator can intervene via teleoperation tools, such as a SpaceMouse.
The learner process samples data evenly from two replay buffers and updates the policy using RLPD. When gripper control is required, a grasp critic is additionally trained with DQN. }
    \label{fig:method}
\end{figure}

Our system is composed of three major components: the actor process, the learner process, and the replay buffer residing inside the learner process, all operating in a distributed fashion, as illustrated in Fig.~\ref{fig:method}. 
The actor process interacts with the environment by executing the current policy on the robot and sends the data back to the replay buffer. 
The environment is designed to be modular, allowing for flexible configuration of various devices. This includes support for multiple cameras, integration of input devices like SpaceMouse for teleoperation, and the ability to control a variable number of robot arms with different type of controllers. An implemented reward function is required to assess the success of a task, which is trained offline using human demonstrations.
Inside the actor process, a human can intervene the robot by using a SpaceMouse, which will then take over the control of the robot from the RL policy.
We employ two replay buffers, one to store offline human demonstrations, called the demo buffer, usually on the range of 20-30; the other one for storing the on-policy data, called the RL buffer.

The learner process samples data equally from the demo and RL replay buffers, optimizes the policy using RLPD, and periodically sends the updated policy to the actor process.  In the remainder of this section, we will provide details about the design choices we made for each component.

\subsection{System Design Choices}
The sample efficiency of the proposed system is crucial, as each minute of training to acquire data incurs a cost. Therefore, the training time must remain within a practical range, particularly when handling high-dimensional inputs like images. 
Additionally, the downstream robotic system must accommodate the RL policy so to ensure a smooth and efficient learning process. For example, the actual low-level robot controller would require great care, particularly for most of the precise contact-rich tasks where the robot physically interacts with objects in the environment. Not only does this controller need to be accurate, but it must also be safe enough that the RL algorithm can explore with random actions during training. Thus, to develop such a system capable of performing sample-efficient policy learning in the real world, we made following design choices.

\paragraph{Pretrained Vision Backbones}To facilitate the efficiency of the training process, we utilize pretrained vision backbones to process image data. While this approach is now a common practice in computer vision for the purpose of robustness and generalization~\citep{clip, vit, bit}; in RL, this treatment offers additional benefits, such as optimization stability and exploration efficiency~\citep{yang2019sampleoptimalq, du2020representation}, making this approach particularly advantageous for real-world robotic RL training.  
Our neural network architecture, illustrated in Fig.~\ref{fig:method}, processes multiple images from cameras using the same pretrained vision backbone. Specifically, we utilize a ResNet-10 model~\citep{he2015deep}, pretrained on ImageNet~\citep{deng2009imagenet}, to generate output embeddings. 
These embeddings are then concatenated and further integrated with processed proprioceptive information, facilitating a more efficient and effective learning process.

\paragraph{Reward Function}One crucial aspect of a reinforcement learning system is the reward function, which is used to guide the learning process and determine the quality of the policy. While there are prior works on utilizing reward shaping to accelerate the learning process~\citep{ng1999rewardshaping, florensa2018automatic,florensa17a}, this procedure tends to be task-specific and time-consuming to design. In some complex tasks, it's simply not feasible to perform such reward shaping. We found that using a sparse reward function, alongside human demonstrations and corrections, offers a straightforward and effective setup for a variety of tasks. 
Specifically, we collect offline data and train a binary classifier for each task, which only grants a positive reward upon task completion and zero otherwise.  

%%SL.10.16: I wonder if we can have a more top-down structure to this section, so that there is an actual overview at the top that maps out the ideas, and then subsections to go into detail about these ideas (like this paragraph)
\paragraph{Downstream Robotic System}To accommodate the policy learning process, we made a few particularly important design choices to the downstream robotic system. 
To facilitate spatial generalization, we represent the robot's proprioceptive state in a relative coordinate system, allowing for an ego-centric formulation. Essentially, at the beginning of each training episode, the pose of the robot's end-effector was randomized uniformly within a pre-defined area in the workspace. 
The robot's proprioceptive information is expressed with respect to the frame of the end-effector's initial pose; the action output from the policy is relative to the current end-effector frame. This procedure simulates physically moving the target when viewed relatively from the frame attached to the end-effector. As a result, the policy can succeed even if the object moves or, as in some of our experiments, is perturbed in the middle of the episode. 
For tasks involving dealing with contact, we use an impedance controller with reference limiting in the real-time layer to ensure safety as in~\citep{luo2024serl}. 
For dynamic tasks, we directly command feedforward wrenches in the end-effector frame to accelerate the robot arm, while it doesn't perform closed-loop control around acceleration, we found this simple open-loop control to be sufficient for considered tasks.
For details regarding the observation representation as well as the controller design, please refer to the supplementary material. 

\paragraph{Gripper Control}For tasks involving the control of grippers, we employ a separate critic network to evaluate discrete grasping actions. 
Although this approach might initially seem like an additional overhead or somewhat unconventional, it has proven to be highly effective in practice, particularly when combined with human demonstrations and corrections. The discrete nature of gripper actions in our tasks makes approximating them with continuous distributions more challenging, particularly in the complex tasks considered in this paper. By using discrete actions, we simplify the training process and improve the overall effectiveness of the reinforcement learning system.
Specifically, we solve two separate MDPs in these tasks, $\mdp_1 = \{\states, \actions_1, \rho_1, \transition_1, \reward, \gamma\}$ and $\mdp_2 = \{\states, \actions_2, \rho_2, \transition_2, \reward, \gamma\}$, where $\actions_1$ and $\actions_2$ are the continuous and discrete action spaces, respectively. They both take in the same state observations from the environment such as images, proprioception, gripper status and so forth. The discrete action space $\actions_2$ consists of all possible discrete actions. For a single gripper, these actions are ``open", ``close", and ``stay". If two grippers are involved, the action space expands to $3^2 = 9$ combinations, accounting for all possible actions each gripper can take. The critic update for $\mdp_2$ follows standard DQN practice~\citep{mnih2013playing} with an additional target network to stabilize training as following: 
\begin{align}
    \mathcal{L}(\theta) = \mathbb{E}_{\bs,\ba,\bs'} \left[ \left( r + \gamma  Q_{\theta^{\prime}}(\bs', \argmax_{\ba^{\prime}} Q_{\theta}(\bs', \ba')) - Q_{\theta}(\bs, \ba) \right)^2 \right],
\end{align}
where $\theta'$ is the target network, which can be obtained by Polyak averaging with current network parameters~\citep{vanhasselt2015double}. 
At training or inference time, we first query the continuous actions from the policy in $\mdp_1$, and then query the discrete actions from the critic in $\mdp_2$ by taking the argmax over the critic's output; we then apply the concatenated actions to the robot. 

\subsection{Human-in-the-Loop Reinforcement Learning} 
With the system-level design choices in place, we now describe the human-in-the-loop procedure that we use to accelerate the learning process.
It is well established from the RL theory literature that the sample complexity of learning an optimal policy is closely tied to the cardinality of the state and action spaces as well as the task horizon~\citep{jin2018qlearning,jin2020qlearning,azar2012samplecomplexity,kearns1998polynominal}, assuming an appropriate exploration policy. These factors collectively serve as proxies for the ``upper bound" on the complexity of tasks that can be feasibly solved. 
Specifically, increases in the size of the state/action spaces, task horizon, or their combinations lead to a proportional rise in the number of samples required to learn an optimal policy; eventually crossing a threshold where real-world training of RL policies becomes impractical.

To tackle this challenge in real-world robotics RL training, we incorporate human-in-the-loop feedback to guide the learning process to help the policy explore more efficiently.
Specifically, a human operator supervises the robot during training and provides corrective actions when necessary, as illustrated in Fig.~\ref{fig:method}.
For an autonomous rollout trajectory from time step $t_0$ to $t_N$, a human can intervene at any time step $t_i$ where $t_0 \leq t_i < t_N$. During an intervention, the human takes control of the robot for up to $N$ steps. Multiple interventions can occur within a single trajectory, as illustrated by the red segments in Fig.~\ref{fig:method}. When a human intervenes, their action $a_{itv}$ is applied to the robot instead of the policy's action $a_{RL}$.
We store the intervention data in both the demonstration and RL data buffers. However, we add the policy's transitions (i.e., the states and actions before and after the intervention) only to the RL buffer. This approach has proven effective in enhancing policy training efficiency.

This intervention is crucial in scenarios where the policy leads the robot to an unrecoverable or undesirable state, or when it becomes stuck in a local optimum that would otherwise require a significant amount of time to overcome without human assistance. 
This procedure is similar to that of HG-DAgger~\citep{Kelly2018HGDAgger}, where a human takes over the control of the robot to collect data when the policy is performing poorly; but our approach uses these data to optimize the policy with reinforcement learning rather than supervised learning, similar to~\citet{luo2023rlif}. In our setup, the human operator engages with a SpaceMouse 3D mouse to provide corrective actions to the robot. 

In the beginning of the training process, the human intervenes more frequently to provide corrective actions, gradually decreasing the frequency as the policy improves. In our experience, we note that the policy improves faster when the human operator issues specific corrections while letting the robot explore on its own otherwise. 

\begin{figure}
    \centering
    \includegraphics[width=1\linewidth]{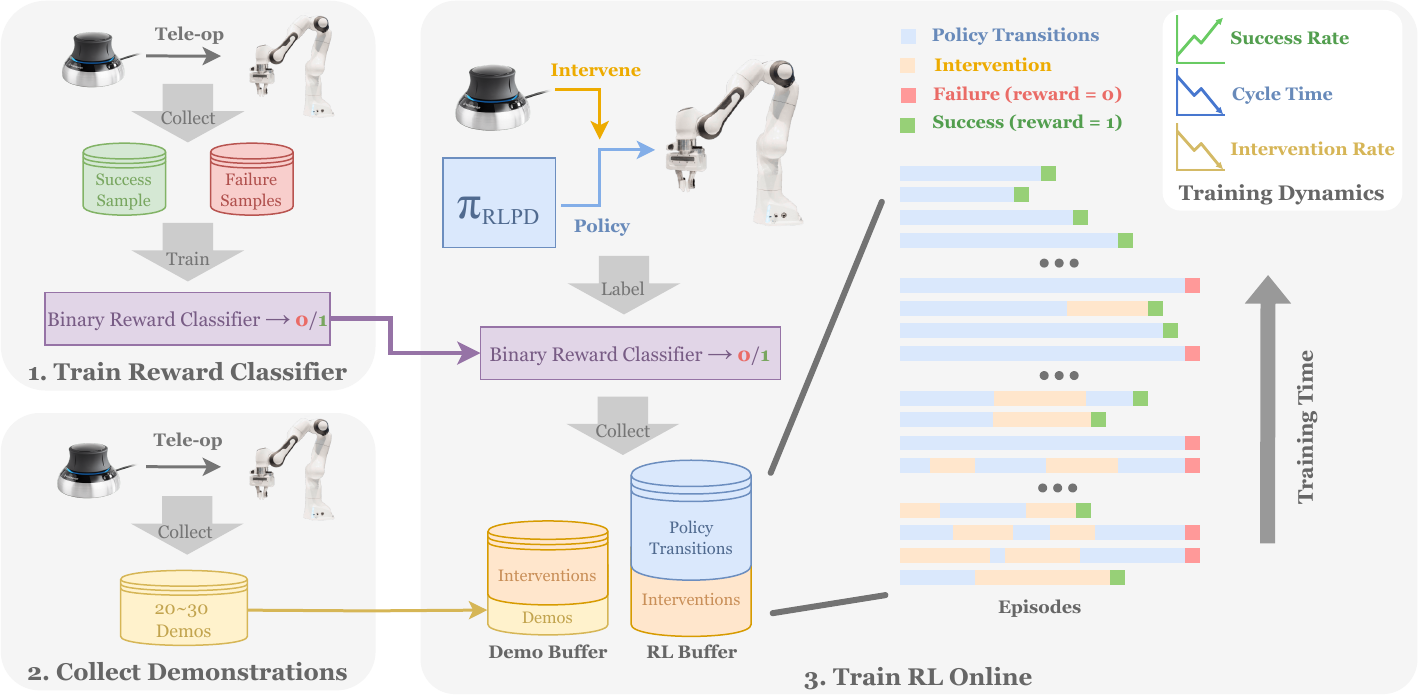}
    \caption{This diagram illustrates the process for training HIL-SERL. First, we tele-operate the robot to collect positive and negative samples and train a binary reward classifier. We then collect a small set of demonstrations, which is added to the demo buffer at the start of RL training. During online training, we use the binary classifier as a sparse reward signal and provide human interventions. Initially, we provide more frequent interventions to demonstrate ways of solving the task from various states. We gradually reduce the amount of interventions as the policy reaches higher success rate and faster cycle times. }
    \label{fig:training-process}
\end{figure}

\subsection{Training Process}
To articulate the training process of our system and assist readers in reproducing our results, we provide a detailed walkthrough of the steps involved in each of our experiments, visually illustrated in Fig.~\ref{fig:training-process} as well.

First, we select cameras that are most suitable for the task. Wrist cameras are particularly useful for facilitating the spatial generalization of the learned policy due to the ego-centric views they provide. However, if wrist cameras alone cannot provide a full view of the environment, we also place several side cameras. 
For all cameras, we perform image cropping to focus on the area of interest and resize the images to 128x128 for the neural network to process.

Next, we collect data to train the reward classifier, which is a crucial step in defining the reward function that guides the learning process. Typically, we gather 200 positive data points and 1000 negative data points by tele-operating the robot to perform the task. This is approximately equivalent to 10 human trajectories, assuming each trajectory takes about 10 seconds. Using our data collection pipeline, as detailed in the supplementary code, it usually takes around 5 minutes to collect these data points.
Additionally, we may collect extra data to address any false negative and false positive issues with the reward classifier. 
The trained reward classifier generally achieves an accuracy of greater than 95\% in the evaluation data set.

We then collect 20-30 trajectories of human demonstrations solving the tasks and use them to initialize the offline demo replay buffer. For each task, we either script a robot reset motion or let the human operator manually reset the task at the beginning of each trajectory, such as the USB pick-insertion task. Finally, we start the policy training process. During this phase, human interventions may be provided to the policy if necessary, until the policy converges. It's also important to note that we should avoid persistently providing long sparse interventions that lead to task successes. Such an intervention strategy will cause the overestimation of the value function, particularly in the early stages of the training process; which can result in unstable training dynamics.

\section{Experiment Results}\label{sec:results}
In this section, we discuss our experiments. We first present the experimental setup and the results. We then discuss these results and their implications.

\subsection{Overview of Experiments}
We conduct experiments across seven diverse tasks covering a range of distinct characteristics, as illustrated in Fig.~\ref{fig:task_filmstrip}. 
These tasks encompass a range of manipulation challenges, including dynamic object manipulation (e.g., flipping an object in a pan), precise and delicate manipulation (e.g., inserting an SSD into its matching slot), combined dynamic and precise manipulation (e.g., inserting a component while the target is moving), flexible object manipulation (e.g., assembling a timing belt) and multi-stage tasks with multiple sub-tasks (e.g., assembling an IKEA shelf). We solve these tasks by utilizing either a single robot arm or a dual-arm setup, together with various combinations of observations and actions.

The observation space can include images from wrist-mounted and side cameras, end-effector poses, twists, forces/torques, and the current gripper status of both arms. 
For dynamic tasks, the action space directly commands feedforward wrenches in the end-effector frame, which can be roughly thought of as desired accelerations.

For other tasks, the action space can include each arm's 6D Cartesian twist target for the downstream impedance controller, and discrete actions for controlling one or two grippers.

For all tasks, unless otherwise noted, we trained a binary classifier as reward detector, it takes images from wrist and/or side cameras as inputs, and predicts whether the current state is a success or failure completing the current task. 
To train such classifiers, we collect both positive and negative demonstrations from human operators, we also collect additional potential false positive or false negative examples if necessary. We include details of the training process for each task in the supplementary material.
For tasks involving grasping, we also include a small negative penalty for gripper actions to discourage the policy from operating its grippers unnecessarily. 
Each task also uses either a scripted robot motion or manually human reset to randomize the initial state of the task. 
Details for setup of each task and policy training can be found in the supplementary material. 
In the remainder of this section, we will first describe each task in detail, and present relevant results as well as comparisons to other state-of-the-art methods.

\subsection{Description of Tasks}
\begin{figure}[t!]
    \centering
    \includegraphics[width=\textwidth]{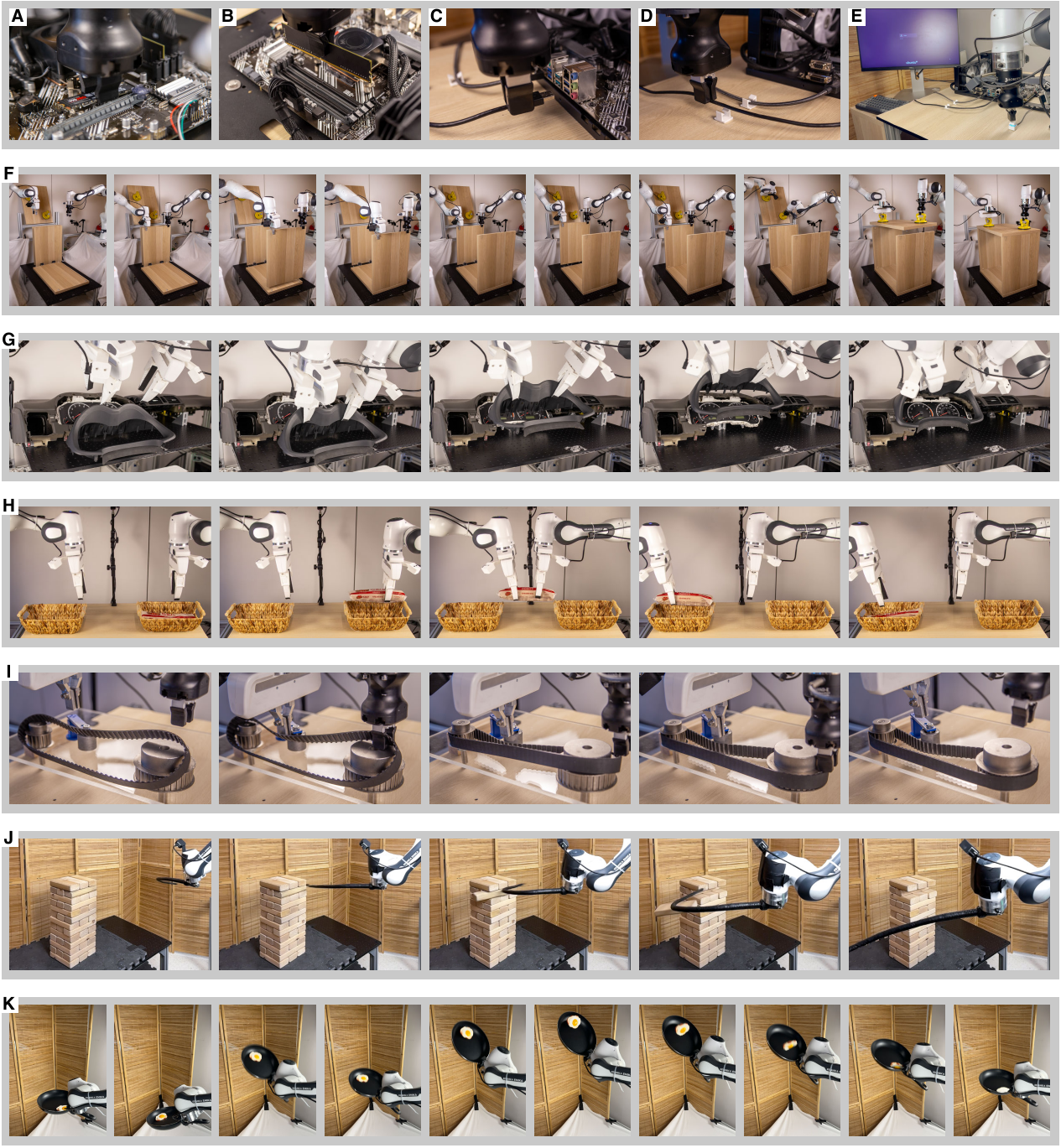}
    \caption{\textbf{Illustrations of the tasks in our experiments.} \textbf{(A)-(E)} A sequence of motherboard assembly tasks: SSD installation, RAM insertion, USB cable grasping and insertion into a slot and a clip, and booting up the computer to ensure motherboard functionality. \textbf{(F)} A manipulation sequence to assemble an IKEA furniture: the robot first assembles two side panels, then installs the top panel onto the mounted side panels. \textbf{(G)} A manipulation sequence to assemble a car dashboard, two robot arms first grasp the workpiece then align multiple pins to the slots. \textbf{(H)} Two arms performing a coordinated handover task. \textbf{(I)} Two arms performing a timing belt installation task. \textbf{(J)} A manipulation sequence of Jenga whipping task, where the robot needs to extract one Jenga piece from the tower without crashing it. \textbf{(K)} The robot is flipping the object in the pan to the opposite side.
    } 
    \label{fig:task_filmstrip}
\end{figure}

In this subsection, we will present descriptions of the tasks in our experiments. We pick our tasks to cover broad range of manipulation challenges, including contact-rich dynamics, dual-arm coordination, flexible object handling, and dynamic manipulation. Here we organize the tasks in a way that similar challenges are presented together. We first present two tasks that require precise manipulation in a contact-rich setting, followed by three tasks that require dual-arm coordination to solve hard tasks including flexible object manipulation. We then proceed to two tasks that require dynamic manipulation. An illustration of each task can be found in Fig.~\ref{fig:task_filmstrip}.

\paragraph{Motherboard Assembly}
The motherboard assembly task includes four subtasks: inserting a RAM card into its matching slot, assembling a PCI-E SSD into the motherboard, picking up a free-floating USB cable and inserting it into the slot, and securing the USB cable into a tight-fitting clip. 

\subparagraph{RAM Insertion} In this task, the robot is supposed to insert a RAM card into the matching slot. The process involves two main steps: it first needs to align the RAM card with the narrow openings on both sides of the slot, then proceed with delicate downward motion with appropriate force to insert the RAM card into the slot. The task is considered successful if the RAM card is fully inserted into the slot without triggering the locking mechanism, allowing for easy reset. If desired, an additional downward force can be applied after executing the trained policy to lock the RAM card in place.
This task is challenging because applying slightly excessive force can cause the RAM card to tilt within the gripper, leading to failure, while insufficient force may result in the RAM card not being properly inserted into the slot. The RAM card is assumed to be pre-grasped by the robot, though we also periodically place it back to a fixture and regrasp it to introduce grasping variations.

\subparagraph{SSD Assembly} In this task, the robot is supposed to insert one side of the SSD into its matching slot and then place the other side onto the fixture residing in the motherboard. The task is considered successful if both sides of the SSD are properly mated into their counterparts. This task requires a gentle but precise insertion strategy initially to avoid damaging the contact pins, followed by another precise motion to align the other side with the supporting fixture. The SSD is assumed to be pre-grasped by the robot, though we also periodically place it back to a fixture and regrasp it to introduce grasping variations.

\subparagraph{USB Connector Grasp-Insertion} In this task, a USB cable is freely placed on a table, and the robot is supposed to grasp the USB connector part, insert it into the corresponding slot and release the gripper. This task is considered successful if the USB connector is fully inserted into the slot and the gripper is released. The difficulty lies in the variability in the initial placement of the USB cable, as does the uncertainty in the grasping pose; the policy must learn to account for such uncertainty during insertion. For example, if an unsuitable grasp was executed,  the policy might need to release the object and regrasp it to achieve a better grasp pose.

\subparagraph{USB Cable Clipping} This task assumes a USB cable is already plugged into the motherboard, and the robot is tasked to pick up the remaining part of the cable and insert it into a tight-fitting organization clip. The task is considered successful if the USB cable is fully inserted into the clip. The difficulty lies in the variability of the deformable USB cable, as well as the tight insertion phase.

\subparagraph{The Whole Assembly} We also performed the whole assembly task by chaining the above four subtasks together, using scripted motions to transition between subtasks. A video clip of the entire assembly process can be found on our project website as well as in the supplementary material. The video demonstrates that the computer successfully booted up after executing the whole assembly policy, validating the effectiveness of our method in not only achieving task success but also performing the RL training process gracefully without damaging these delicate components.

\paragraph{IKEA Assembly}
The IKEA assembly task involves assembling an IKEA shelf with four boards, and is decomposed into three subtasks: the robot needs to first assemble the two side panels into a panel fixed on the table, then mount the top panels into side panels after they are assembled. The task is considered successful if all the pieces are properly assembled into the shelf. For all the subtasks, we assume the panels are pre-grasped by the robot, however, we do periodically place them back to fixtures and regrasp them to introduce grasping variations.

\subparagraph{Side Panel Assembly} In this task, the top part of the side panel is assumed to be pre-grasped by the robot, though the grasping location can vary due to the heavy weight of the panel during the interactive assembly process, and we do periodically regrasp them from the fixtures. The task is considered successful if the bottom part is properly assembled onto the two matching pins.

\subparagraph{Top Panel Assembly} After two side panels are assembled, this task demands the robot to mount the top panels onto both of the side panels. The task is considered successful if all four pins of the top panel are properly inserted into the corresponding holes on the side panel. This task is difficult because the top parts of the side panels can move around during the assembly process, and the policy must adapt to these variations to succeed. 

\subparagraph{The Whole Assembly} We also performed the whole assembly task by chaining the three trained policies, using scripted motions to transition between subtasks, which is illustrated in Fig.~\ref{fig:task_filmstrip} and the supplementary material. For each subtask, we uniformly randomized the scripted grasping poses by 1 cm in each translation dimension to introduce variations to the policy. This task is considered successful if all the panels are assembled successfully, and each sub-policy is allowed for a maximum of two attempts. For this task, we perform 10 trials for the chained policies, since it's a very long-horizon task.

\paragraph{Car Dashboard Assembly} As illustrated in Fig.~\ref{fig:task_filmstrip}, the car dashboard assembly task involves two stages: both of the arms first need to grasp on appropriate locations of the workpiece and then lift it up, assemble it onto the dashboard. The task is considered successful if all the pins of the workpiece are fully inserted into the corresponding holes on the dashboard. This task requires precise manipulation as well as bimanual coordination: the two arms must coordinate the timing of the motions and gripper closure to lift the workpiece up, rotate it and align multiple pins at the same time.

\paragraph{Object Handover} In this task, two robot arms are required to coordinate transferring an object from one basket to the other. The right arm first picks up the object from a basket on the right side. Then, it hands the object over to the left arm, which places it precisely into a basket on the left side. The task is considered successful if the object is placed flat on the right basket. The handover part is challenging because the robots' grippers must coordinate the timing of the motions to prevent the object from falling down.

\paragraph{Timing Belt Assembly} In this task, two robot arms collaborate to assemble a timing belt onto pulleys and actuate the tensioner. This task is part of the NIST board assembly challenge~\citep{nistboard}. The process involves locating and manipulating the randomly placed belt, coordinating precise motions to thread the belt over two pulleys, and simultaneously actuating the tensioner to accommodate the belt. Success is achieved when the belt is properly threaded onto both pulleys and the tensioner is securely tightened.
This task presents several challenges. The belt can deform unpredictably during the assembly process, requiring adaptive manipulation. The arms must coordinate precisely, with carefully timed movements to thread the belt effectively. The timing of tensioner adjustments is critical, as the tensioner must allow the belt to be threaded while avoiding jamming. Throughout the process, the policy must continuously adjust to the changing state of the flexible belt and the overall system configuration.
The complexity of this task stems from the need to handle a deformable object while precisely coordinating bimanual actions and managing the tensioner mechanism. This requires the policy to develop sophisticated, reactive behaviors to succeed consistently.

\paragraph{Jenga Whipping} In this task, the robot is supposed to whip out a specific block from a Jenga tower without toppling over the tower. The nature of this task is largely different from the previous tasks, in that it requires the robot to learn a highly dynamic open-loop behavior, as opposed to the reactive closed-loop behavior required in the previous tasks. The dynamics of this task are intractably complex: the deformable whip travels at a very high speed and interacts with the surrounding compressed air, making its motion difficult to predict. Additionally, determining the precise force needed to remove a specific block without destabilizing the entire tower introduces further complexity due to the intricate contact dynamics involved. It is imperative for the policy to develop a reflex-like behavior by observing the outcomes of its own actions, intuitive physics, and the interactions between the whip and the blocks. This allows the robot to execute precise and consistent motions to successfully remove the target block without causing the tower to collapse. Note for this specific task, we initialize an offline dataset with 30 expert demonstrations rather than using real-time human corrections. This was chosen deliberately, as incorporating human feedback during training would be both impractical and unsuitable given the task's unique characteristics.

\paragraph{Object Flipping} In this task, an object is randomly placed on a pan attached to the robot's end-effector, and the robot is tasked to flip the object over a horizontal axis. The task is considered successful if the object is flipped to the opposite side and remains in the pan. Since the initial placement of the object is randomized, the policy must learn to adapt to these variations, e.g., moving the object to a more favorable position before executing the flip motion. The task's nature is similar to the Jenga task, requiring precise and sophisticated open-loop behaviors. However, it also involves a closed-loop component, as the policy might need to reposition the object initially.

\subsection{Experimental Results}

\begin{table}[ht!]
\centering
\small

\begin{subtable}{\linewidth}
\centering
\setlength{\tabcolsep}{4pt} % Reduce column separation
\renewcommand{\arraystretch}{1.2} % Reduce the vertical spacing

\begin{tabularx}{\linewidth}{@{}r|>{\centering\arraybackslash}p{0.09\linewidth} |>{\centering\arraybackslash}p{0.11\linewidth} Y|>{\centering\arraybackslash}p{0.11\linewidth} Y@{}}
& \multirow{2}{*}{\textbf{\makecell{Training \\Time (h)}}} & \multicolumn{2}{c|}{\textbf{Success Rate (\%)}} & \multicolumn{2}{c}{\textbf{Cycle Time (s)}} \\
\cline{3-6}
 \textbf{Task} & & BC & HIL-SERL (ours) & BC & HIL-SERL (ours) \\
\hline
RAM Insertion & 1.5 & 29 & \textbf{100} (+245\%) & 8.3 & \textbf{4.8} (1.7x faster) \\
SSD Assembly & 1 & 79 & \textbf{100} (+27\%) & 6.7 & \textbf{3.3} (2x faster) \\
USB Grasp-Insertion & 2.5 & 26 & \textbf{100} (+285\%) & 13.4 & \textbf{6.7} (2x faster) \\
Cable Clipping & 1.25 & 95 & \textbf{100} (+5\%) & 7.2 & \textbf{4.2} (1.7x faster) \\
IKEA - Side Panel 1 & 2 & 77 & \textbf{100} (+30\%) & 6.5 & \textbf{2.7} (2.4x faster) \\
IKEA - Side Panel 2 & 1.75 & 79 & \textbf{100} (+27\%) & 5.0 & \textbf{2.4} (2.1x faster) \\
IKEA - Top Panel & 1 & 35 & \textbf{100} (+186\%) & 8.9 & \textbf{2.4} (3.7x faster) \\
IKEA - Whole Assembly & -- & 1/10 & \textbf{10/10} (+900\%) & -- & -- \\
Car Dashboard Assembly & 2 & 41 & \textbf{100} (+144\%) & 20.3 & \textbf{8.8} (2.3x faster) \\
Object Handover & 2.5 & 79 & \textbf{100} (+27\%) & 16.1 & \textbf{13.6} (1.2x faster) \\
Timing Belt Assembly & 6 & 2 & \textbf{100} (+4900\%) & 9.1 & \textbf{7.2} (1.3x faster) \\
Jenga Whipping & 1.25 & 8 & \textbf{100} (+1150\%) & -- & -- \\
Object Flipping & 1 & 46 & \textbf{100} (+117\%) & 3.9 & \textbf{3.8} (1.03x faster) \\
\hline
\textbf{Average} & -- & 49.7 & \textbf{100} (+101\%) & 9.6 & \textbf{5.4} (1.8x faster) \\
\end{tabularx}
\caption{Comparison of BC and RL success rates and cycle times for various tasks. 
All metrics were reported over 100 trials per task, except for the IKEA whole assembly task, which involved 10 trials.
For all tasks, BC baselines were trained using HG-DAgger with the same number of episodes and interventions as RL. However, for the Jenga whipping and object flipping tasks, we used ``flat" BC, trained on 50 and 200 demonstrations, respectively.
}
\label{subtab:tasks}
\end{subtable}

\vspace{1em} % Add some vertical space between subtables

\begin{subtable}{\linewidth}
\centering
\small  % This will make the entire table content smaller
\setlength{\tabcolsep}{4pt} % Reduce column separation
\renewcommand{\arraystretch}{1.2} % Reduce the vertical spacing

\begin{tabular}{l|cccccccccc}
Task & DP & HG-DAgger & BC  & IBRL & Residual RL & DAPG & \multicolumn{1}{c}{\footnotesize$\stackrel{\text{HIL-SERL}}{\text{no demo no itv}}$} & \multicolumn{1}{c}{\footnotesize$\stackrel{\text{HIL-SERL}}{\text{no itv}}$} &  HIL-SERL (ours) \\
\hline
RAM Insertion & 27 & 29 & 12 & 75 & 0 &  8 & 0 & 48 & \textbf{100} \\
Dashboard Assembly & 18 & 41 & 35 & 0 & 0 & 18 & 0 & 0 & \textbf{100} \\
Object Flipping & 56 & 46 & 46 & 95 & 97 & 72 & 0 & \textbf{100} & \textbf{100} \\
\hline
\textbf{Average} & 34 & 39 & 31 & 57 & 32 & 33 & 0 & 49 & \textbf{100}
\end{tabular}
\caption{
Comparison of various methods on selected tasks. Diffusion Policy (DP) and BC are trained with 200 demonstrations, while HG-DAgger is trained with the same number of episodes and interventions as RL. IBRL, Residual RL, and DAPG are initialized with 200 demonstrations. Our method is also ablated with two versions: one initialized from scratch without demonstrations or corrections, and another initialized from demonstrations but without corrections.
}
\label{subtab:baselines}
\end{subtable}

\caption{Experiment results. \textbf{(a)} HIL-SERL against imitation learning baselines. \textbf{(b)} HIL-SERL against various other baselines.}
\label{tab:results}
\end{table}

\begin{figure}[t!]
    \centering
    \includegraphics[width=\textwidth]{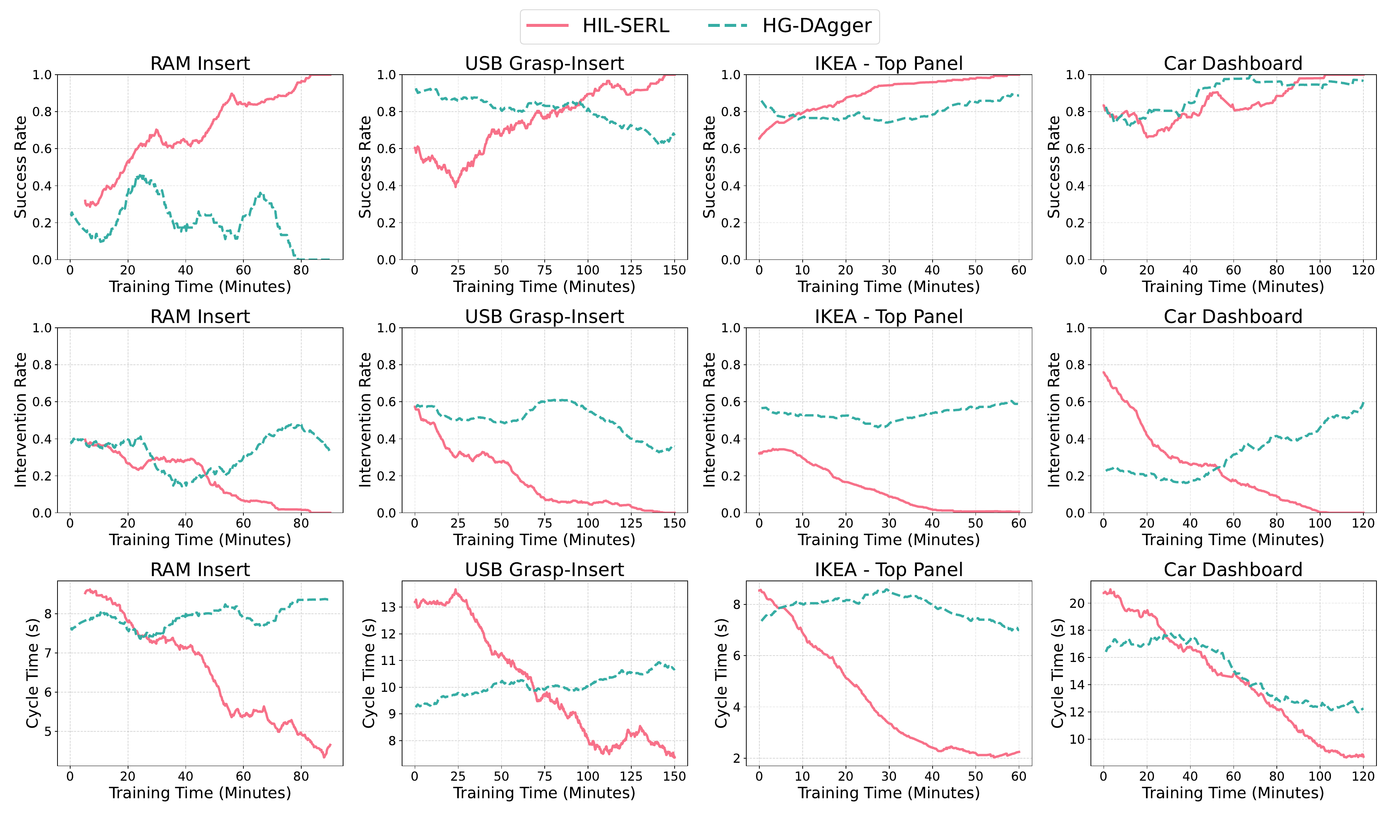}
    \caption{\textbf{Learning curves for experimental tasks.} This figure presents the success rate, cycle time, and intervention rates for both HIL-SERL and DAgger across few representative tasks, displayed as a running average over 20 episodes. 
    For HIL-SERL, the success rate increased rapidly throughout training, eventually reaching 100\%, while the intervention rate and cycle time progressively decreased, with the intervention rate ultimately reaching 0\%. 
    For HG-DAgger, the success rate fluctuates throughout training episodes and does not necessarily increase as training progresses. Since interventions occur frequently, leading to successful outcomes, the true policy success rate is likely lower than the curve suggests. Additionally, the intervention rate does not consistently decrease over time, indicating that the policy is not steadily improving. This is reflected in the cycle time as well, which shows no improvement, as DAgger lacks a mechanism to enhance performance beyond the provided training data. Additional plots are available in the supplementary material.
    }
    \label{fig:task_results}
\end{figure}

In this subsection, we present the experimental results for all the tasks mentioned above. For each task, we report the success rate, cycle time, and training time. 
The training time includes all scripted motion, policy rollouts, intended stops, as well as onboard computation which is carried on a single NVIDIA RTX 4090 GPU.
Unless otherwise noted, all results are based on 100 evaluation trials. During these evaluations, we randomize the initial states using either scripted robot motions or human resets. Our evaluation protocol can be found in the supplementary material.

A central claim of this paper is that HIL-SERL outperforms imitation learning methods based on human teleoperation. To substantiate this, it is crucial to compare relevant imitation learning methods fairly under equivalent settings. Na\"{i}ve imitation learning approaches often suffer from error compounding issues, as noted by \citep{dagger}. DAgger and its variants \citep{dagger, Kelly2018HGDAgger} address this problem by incorporating human corrections to refine the policy through supervised learning.
Our method also leverages human corrections, but instead utilizes them to optimize the policy through reinforcement learning based on task-specific rewards. Therefore, we compare our approach to imitation learning by training a baseline with HG-DAgger~\citep{Kelly2018HGDAgger},
using the same amount of human demonstrations and corrections as used in reinforcement learning. We first pretrain a base policy with behavioral cloning (BC) using an equivalent amount of offline human demonstrations as provided to our method. We then run this policy and collect human expert corrections, such that the total amount of trials and interventions matches RL training. Specifically, we run it for the same number of episodes as our method and aim to provide a comparable number of interventions per episode.

This comparison is performed for all tasks except Jenga whipping and object flipping, where interventions are challenging and undesirable. For these tasks, we instead collect 50 and 200 offline demonstrations and train BC policies as baselines. This provides a significantly larger number of demonstrations than our method, which typically requires only 20-30 demonstrations.

In all our experiments, we use success rates and cycle time as primary metrics to compare different methods. To further validate the effectiveness of our approach, we also report the intervention rate over time, demonstrating that our policy improves progressively, reducing the need for interventions. Ideally, the intervention rate trends toward zero, indicating that the policy performs autonomously.
The experiment results can be found in Fig.~\ref{fig:task_results} and Table.~\ref{subtab:tasks}.

First, as shown in Table.~\ref{tab:results}, HIL-SERL achieved a success rate of 100\% within 1 to 2.5 hours of real-world training on nearly all the tasks. This is a significant improvement over the HG-DAgger baseline, which achieved an average success rate of 49.7\% across all tasks. The performance gap is more pronounced for the tasks that require more complex behaviors -- Jenga whipping, RAM stick insertion, and timing belt assembly.

We also report the number of human interventions over time for nearly all of the tasks in Fig.~\ref{fig:task_results}. Specifically, we report the intervention rate, for which we calculate the ratio of intervened timesteps to total timesteps within an episode and report a running average over 20 episodes.
As shown in the figure, the intervention rate decreases as the training progresses, indicating that the policies are improving and less reliant on human corrections. Additionally, we observe that the total duration of interventions decreases dramatically. Initially, we issue long, sparse interventions when the policy is immature. As the policy improves, shorter interventions are sufficient to correct fewer mistakes. In contrast, the HG-DAgger policies require more frequent interventions to correct the policy, and the total duration of interventions does not necessarily decrease over time. Thus, our method attains better performance with less human supervision.

Our method outperforms HG-DAgger due to key advantages of RL. RL explores a wider range of states and directly optimizes task-specific rewards, while DAgger’s reliance on human corrections can introduce inconsistencies and limit state exploration. As RL learns from its own state distribution and corrects errors autonomously, it overcomes the constraints of human demonstrations, resulting in more robust policies.
These empirical findings align with theoretical results discussed in \citet{luo2023rlif}, which demonstrate that RL policies can in principle outperform DAgger. The performance gap tends to widen as the suboptimality of human corrections increases, a scenario that is more likely to occur as tasks become more complex.

Another important aspect to consider is the cycle time, or the time it takes to complete the task. On average, the HG-DAgger policies achieve an average cycle time of 9.6 s, while our method achieves an average cycle time of 5.4 s. This indicates an improvement of 1.8 times faster.
This improvement is expected, as imitation learning methods lack mechanisms to deal with suboptimality in human demonstrations.
In contrast, reinforcement learning (RL) can leverage dynamic programming to optimize for the discounted sum of rewards. For a discount factor 
$\gamma < 1$, this approach encourages the policy to acquire rewards faster, resulting in shorter cycle times compared to those achieved by imitating human demonstrations.

Out of these experiments, we would note that our method proves to be general and effective across tasks with vastly different physical properties, generating both open-loop and closed-loop policies well suited for each task's specific requirements.
For precise manipulation tasks, such as assembling a timing belt or inserting a RAM stick, the policy learns to associate task-relevant visual features with appropriate twist motions. It performs continuous visual servoing behavior, reacting to streaming observations in real-time and adjusting its motion until reaching the target. In contrast, for tasks like Jenga whipping and object flipping, the policy learns to predict potential outcomes of its actions through interaction. It then precisely refines its motion to achieve the desired outcome while maintaining consistency in execution. We also offer an in-depth analysis of the learned behavior, which we defer to the later section.

Overall, our experiments show that our method is both general and effective. It successfully learns policies for a variety of challenging manipulation tasks using the same approach, achieving high performance across all tasks. Additionally, it achieves such performance within practical training times, even for high-dimensional observations and action spaces, such as those required for bimanual manipulation.

\subsection{Robustness Results}

\begin{figure}[t!]
    \centering
    \includegraphics[width=\textwidth]{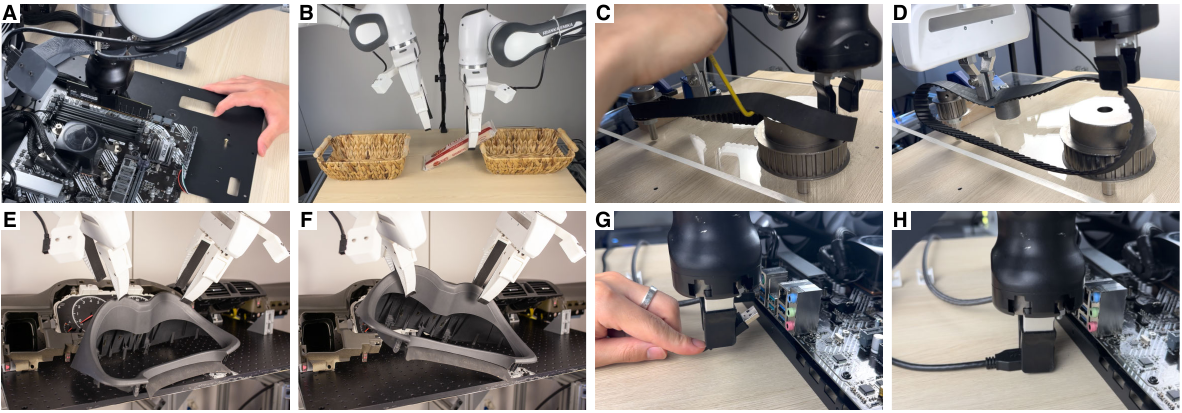}
    \caption{\textbf{Robustness evaluation for policies learned by our method.} \textbf{(A)} RAM insertion under external perturbations, such as a moving motherboard. \textbf{(B)} Retrying behavior during a handover task after the grippers are forced open. \textbf{(C-D)} Reactive responses in the timing belt task, addressing both external disturbances and unexpected deformations during execution. \textbf{(E-F)} In the dashboard assembly task, the policy performs re-grasps after one or both grippers are forcibly opened. \textbf{(G-H)} In the USB grasp-insertion task, the policy adapts to external disturbances and poor grasps by releasing and regrasping the object.
    }
    \label{fig:robustness}
\end{figure}

To test the zero-shot robustness of policies learned by our method, we provide a set of qualitative results in Fig.~\ref{fig:robustness}. 
These results demonstrate the policy's ability to adapt dynamically to variations on the fly and handle external disturbances, such as objects being intentionally dropped by a human from the gripper, or cases where a human deliberately forces the gripper open during task execution.
Corresponding video clips can be found in our supplementary material and website \url{https://hil-serl.github.io/}

In the timing belt assembly task, the belt can undergo arbitrary deformation, and the policy is supposed to adapt to these variations during the assembly process. Additionally, we introduce external disturbances to the belt to further test the robustness of the policy. These disturbances include artificially perturbing the belt's shape or repositioning it dynamically during the assembly process, as in Fig.~\ref{fig:robustness} (C) and (D).
In the RAM insertion task, the learned policy successfully inserts the RAM stick even when the target is moving during the insertion process, thanks to the ego-centric representation of the proprioceptive observation, as illustrated in Fig.~\ref{fig:robustness} (A). 
For the car dashboard assembly and object handover tasks, after the policy grasps the object, we force the gripper to open during task exectution. The policy reacts to this unexpected situation by retrying to grasp the object and proceeding with the rest of the tasks, as presented in Fig.~\ref{fig:robustness} (B), (E) and (F). 
In the USB connector grasp-insertion task, we varied the initial pose of the USB connector and occasionally forced it out of the gripper to simulate poor grasp poses. 
The policy adapted by releasing the connector and regrasping it to achieve a better pose for insertion, as seen in Fig.~\ref{fig:robustness} (G) and (H).
These robust behaviors are achieved through autonomous exploration during the RL training phase. For example, the policy learns to associate the grasping pose with the downstream insertion task and regrasps the object if necessary. However, these robust behaviors are usually hard to achieve with imitation learning methods, as they lack this mechanism to autonomously explore and learn from the outcomes of their actions.

\subsection{Additional Baseline Comparisons}
To validate the effectiveness of design choices in our method, we conducted additional comparisons on three representative tasks: car dashboard panel assembly (dual-arm coordination), RAM insertion (precise manipulation), and object flipping (dynamic manipulation).
We compare our approach against several state-of-the-art methods to highlight different aspects of its performance. To illustrate the significance of human interventions, we performed ablation studies varying the number of human demonstrations and corrections. To showcase how effectively our method incorporates and leverages human demonstrations, we compared against Residual RL~\citep{residualrl}, DAPG~\citep{Rajeswaran-RSS-18}, and IBRL~\citep{hu2024ibrl}. Additionally, we compare against Diffusion Policy~\citep{chi2024diffusion},
to ensure that the task difficulty does not stem solely from multi-modality in human demonstrations. These comprehensive comparisons serve to validate our method's effectiveness and capabilities across diverse manipulation scenarios, results are presented in Table.~\ref{tab:results}.

We first note that RL from scratch, without any demonstrations or corrections, achieved 0\% success rate on all tasks. 
To validate the importance of online human corrections, we increased the number of demonstrations in the offline buffer of SERL tenfold, from the usual 20 to 200. However, without any online corrections, this approach resulted in significantly lower success rates compared to HIL-SERL, including a complete failure (0\% success) on complex tasks such as the car dashboard assembly. This confirms the crucial role of online corrections in facilitating policy learning.
These results confirmed the crucial role of both offline demonstrations and on-policy human interventions in guiding policy learning, especially for complex manipulation tasks that require continuous reactive behaviors.

For the object flipping task, we trained BC policies using both 20 and 200 demonstrations. The results from these two BC policies were quite similar, with success rates of 47\% and 46\%, respectively, even though the number of demonstrations increased tenfold. This indicates that merely imitating human demonstrations is insufficient to solve this task, even though it is largely open-loop. 

Another important aspect to consider is how our method handles demonstrations compared to others. 
To compare against the mentioned baselines, we collected 200 demonstrations for each task. Note this number is substantially larger than the number of offline demonstrations in our method, which is usually around 20-30 in the offline replay buffer. 
For Residual RL and IBRL, we trained behavior cloning (BC) policies with these demonstrations to feed into their algorithm pipeline. For DAPG, we stored these 200 demonstrations in a separate buffer and regularized the policy actions towards them.
Overall, our method consistently outperformed these baselines by large margins, as shown in Table.~\ref{tab:results}. 

This can be interpreted as follows. Residual RL depends on a pre-trained BC base policy to facilitate the learning process. However, this approach can be problematic for tasks that require precise manipulation, such as car dashboard assembly or RAM insertion. In these scenarios, imitation learning methods, including BC, often perform inadequately. As a result, the RL policy learning process can experience significant failures.
For IBRL, the actor is a hybrid of a BC policy and an RL policy, making the behavior more "BC-like." This approach struggles on tasks where BC performs poorly. For DAPG, since it directly regularizes the policy actions towards the demonstrations, it is unsurprising that the policy performs similarly to the BC policies. Therefore, it underperforms our method on tasks that require more reactive and complex behaviors.

The effectiveness of our method comes from the off-policy nature of the underlying RL algorithm, which dynamically weights human data based on its relevance to the current policy optimization objective. Different from ~\citet{residualrl,hu2024ibrl,Rajeswaran-RSS-18} which heavily relies on the quality of human demonstrations, our method has a mechanism allows for efficient use of human data early in training while enabling the agent to progressively surpass human-level performance. Crucially, it prevents the agent from being constrained by the limitations of human demonstrations, striking a balance between bootstrapping from demonstrations and discovering novel, superior strategies through autonomous exploration.

To compare with the Diffusion Policy~\citep{chi2024diffusion}, We trained the policies using 200 demonstrations for each task, which is substantially more than the 20 demonstrations available in the offline replay buffers used by our method. 
We report the results using the experimented optimal algorithm parameters, such as observation and action chunking length, and the length of the applied action sequence in the supplementary material. 
On the RAM insertion and car dashboard panel tasks, diffusion policies achieved success rates of 27\% and 28\%, respectively. On the object flipping task, the success rate is 56\%.
This performance is significantly lower than our method and even falls short of the HG-DAgger baseline. 
This outcome is not surprising, as the primary strength of diffusion policies is in learning a more expressive policy distribution to accurately ``memorize" robot motions. 
However, these tasks require more ``closed-loop" reactive behaviors, such as continuous visual servoing to correct motion errors. Therefore, the advantage of diffusion policies in learning an expressive policy distribution does not necessarily lead to better performance in these tasks.

\section{Result Analysis}\label{sec:analysis}
To provide deeper insights into our results, we offer a detailed analysis of the learned policies. This analysis focuses on two key aspects: reliability and learned behaviors. We examine why the learned policies consistently achieve high success rates across diverse tasks, investigating the factors that contribute to their robustness. 
Additionally, we delve into the nature of the behaviors acquired by the policies, particularly exploring the distinction between reactive and predictive strategies. This comprehensive analysis aims to shed light on the underlying mechanisms that contribute to the effectiveness of our approach in solving complex manipulation tasks.

\begin{figure}[t]
    \centering
    \includegraphics[width=\textwidth]{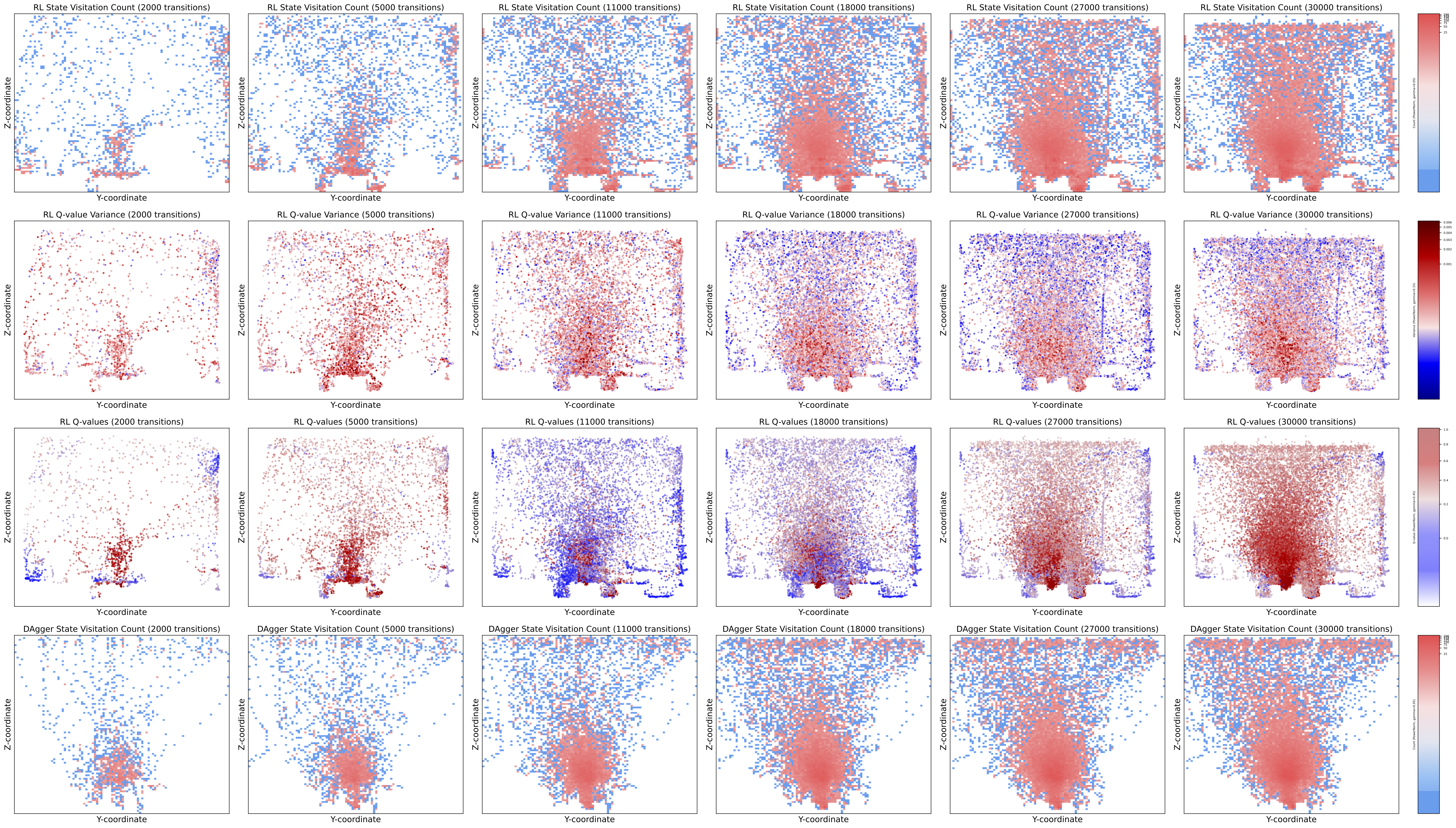}
    \caption{\textbf{Visualization of policy training dynamics.} \textbf{(A)} State visitation heatmaps during HIL-SERL training: The policy progressively forms a ``funnel" shape, concentrating more on areas around the demonstrations and corrections, showing robustification in these regions. \textbf{(B)} Q-value variance scatter plots throughout training: States within the funnel show increased Q-value variance, indicating that the policy is gaining greater confidence in actions that lead to successful outcomes. \textbf{(C)} Q-value scatter plots across training: Critical states, marked by higher Q-value variance, are also associated with high Q-values. \textbf{(D)} State visitation heatmaps during HG-DAgger training: The funnel shape is less pronounced, more diffused distribution of visitation density.}
    \label{fig:analysis-robustness}
\end{figure}

\subsection{Reliability of the Learned Policies}
One key aspect of HIL-SERL's performance is its high reliability, achieving a 100\% success rate across all tasks. 
We argue this reliability comes from reinforcement learning's inherent ability to self-correct through policy sampling, allowing the agent to continuously improve by learning from both successes and failures. In contrast, imitation learning approaches, including interactive methods, lack this self-correction mechanism, making it significantly more challenging to achieve comparable performance with the same amount of data.
While there is existing theoretical work on the convergence of Q-learning~\citep{papavassiliou1999ijcai,bubeck2018qlearning, jin2020qlearning,yang2019sampleoptimalq}, our analysis focuses on providing an intuitive understanding of the training dynamics.

To illustrate this, we analyze the RAM insertion task, which requires precise manipulation and is easily visualized due to symmetrical randomization in the X and Y directions. We plot heatmaps of state visitation counts across timesteps for different policy checkpoints in Fig.~\ref{fig:analysis-robustness}, based on the end-effector's Y and Z positions.
Through policy learning, we observe the gradual development of a funnel-like shape connecting the initial states to the target location. This funnel becomes more defined as empty regions are filled, and narrows when approaching the target, indicating increased policy confidence and precision. Over time, the mass concentrates in areas likely to succeed.
We then introduce the concept of ``critical states", defined as states where the Q-function variance is large. We compute this variance using:
\begin{align}
\text{Var}[Q(\bs, \ba)] &= \mathbb{E}_{{\epsilon \sim \mathcal{U}[-c, c]}} \left[ \left( Q(\bs, \ba + \epsilon) - \mathbb{E}_{{\epsilon \sim \mathcal{U}[-c, c]}}(Q(\bs, \ba + \epsilon)) \right)^2 \right].
\end{align}
For each datapoint and its associated policy checkpoint, we add uniform random noise from [-0.2, 0.2] to the action (normalized to [-1, 1]) at every state and compute the Q-function variance using Monte Carlo sampling with 100 samples. A large variance indicates that the state is critical to the policy's success, as taking a different action would result in significantly different (usually much smaller) Q-values.
Fig.~\ref{fig:analysis-robustness} also shows heatmaps of Q-values and their variances at different states. These plots clearly demonstrate the policy developing a funnel where the most visited states gain higher Q-values and higher Q-value variances. This indicates that the policy is robustifying the region, effectively connecting important states with actions leading to high Q-values through dynamic programming.

In contrast, the heatmap of state visitation counts for HG-DAgger on the same task (fourth row of Fig.~\ref{fig:analysis-robustness}) shows a much sparser distribution. The funnel shape is less distinct, and the mass is more spread out, with states visited more uniformly compared to the RL case. This is because RL can explore autonomously and use dynamic programming directed by task rewards, while DAgger can only explore around the current policy. Consequently, to achieve similar performance, DAgger may require significantly more demonstrations and corrections, as well as careful attention from the human operator to ensure data quality.

This type of stabilizing behavior within a funnel has been studied in state-based dexterous manipulation and motion planning~\citep{burridge1999funnel, tadrake2010lqrtrees}. However, our approach differs in that we directly leverage perceptual inputs and use RL exploration to autonomously form the funnel.
An analogous concept in optimal control is the development of controllers that stabilize around nominal trajectories using local feedback~\citep{murray2008feedback}. In our case, demonstrations and corrections can be regarded as ``nominal trajectories" around which RL methods develop funnels for stabilization.

\subsection{Reactive Policy and Predictive Policy}
\begin{figure}[t!]
    \centering
    \includegraphics[width=\textwidth]{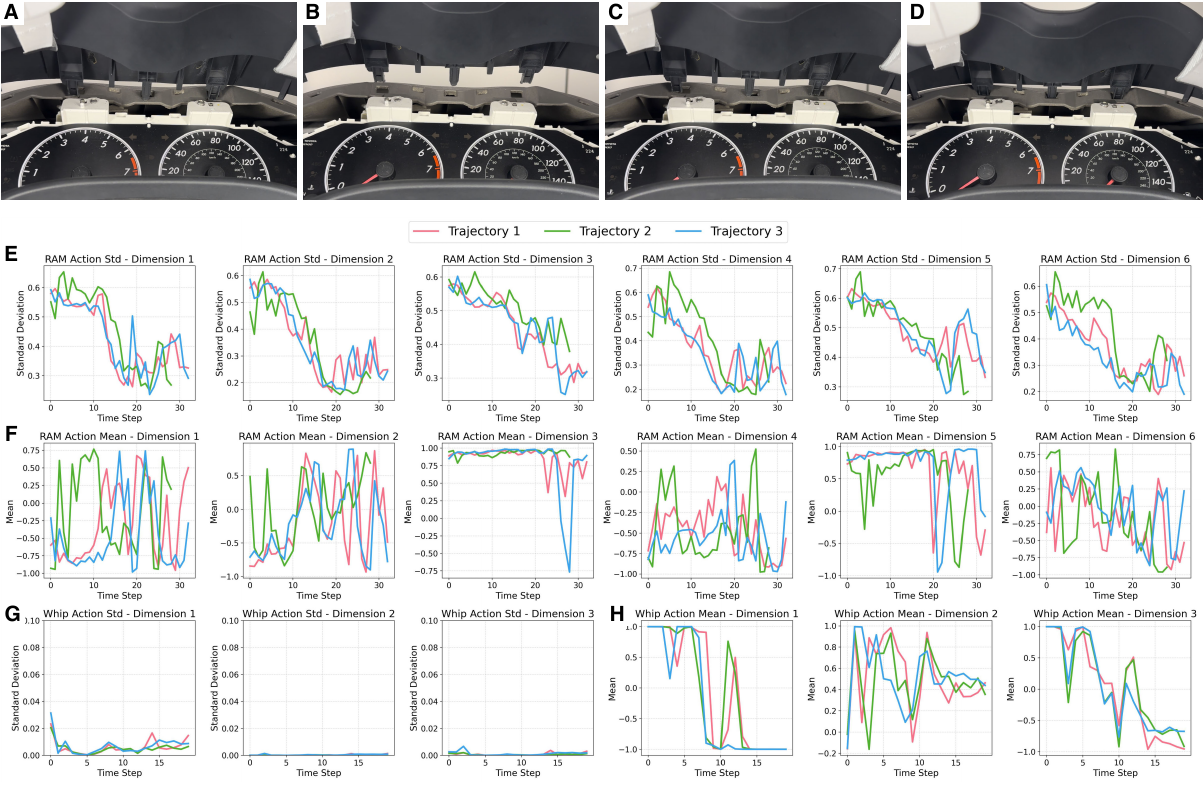}
    \caption{\textbf{Reactive vs Predictive Behavior.}
    \textbf{(A-D)} A sequence of reactive behaviors in the dashboard assembly task: after getting stuck in contact, the policy breaks the contact by quickly lifting two arms, then re-establishing the contact when approaching the target, finally succeeding in the insertion. 
    \textbf{(E)} Variance plots from trained Gaussian policies in the RAM insertion task, showing three trajectories. Initial variance is high but rapidly decreases as the target is approached.
    \textbf{(F)} Mean plots from trained Gaussian policies in the RAM insertion task, with values ranging from -1 to 1.
    \textbf{(G)}  Variance plots in the Jenga whipping task, remaining consistently low (near 0), indicating stable execution and open-loop behavior.
    \textbf{(H)} Mean plots in the Jenga whipping task, with values between -1 and 1, demonstrating consistent behavior across three trajectories.
 }
    \label{fig:analysis-reactive-predictive}
\end{figure}
To solve most of the high-precision manipulation tasks, we need a closed-loop reactive policy that responds rapidly to immediate sensory feedback, therefore enabling precise adjustments in real-time. 
On the other hand, for the dynamic manipulation tasks, such as Jenga whipping and object flipping,
it is desirable to employ an open-loop predictive policy that plans ahead and execute the motion consistently. 
To see this, we pick two representative tasks requiring these two different types of policies, Jenga whipping and RAM insertion, for analysis. 
To visualize the differences between these policy types, we plot the computed actions from the trained Gaussian policies for both tasks in Fig.~\ref{fig:analysis-reactive-predictive}. 

For both tasks, we analyzed three successful trajectories by plotting the policies' computed standard deviation and mean over time. From these plots, we observe that while the mean actions cover a wide range of values in both cases, the standard deviations reveal distinct policy behaviors.
In the Jenga whipping task, the standard deviation remains consistently low (very close to 0) across time steps. This indicates a highly confident and consistent policy, ideal for tasks where open-loop behaviors are desirable. Similar to a tennis player developing a reflex, the policy learns to execute a precise, pre-planned motion. Through environmental interactions, it refines this motion to minimize prediction errors, resulting in consistent execution.
Conversely, the RAM insertion task exhibits a different pattern. Initially, the standard deviation is much higher (around 0.6), reflecting uncertainty when approaching the target early on. However, it decreases rapidly over time, suggesting an initially coarse approaching motion that becomes more precise when near the target. This task demands a reactive policy capable of error correction in various scenarios, as predictive control over a long horizon is impractical due to the task's precision requirements.
This reactive behavior is even more pronounced in complex manipulation tasks such as dashboard panel assembly or timing belt installation. In these cases, the policy must continuously adjust its actions based on sensory feedback, often requiring multiple attempts to achieve success, such as breaking contact and re-approaching the target, as illustrated in Fig.~\ref{fig:analysis-reactive-predictive}. The higher variance in these scenarios indicates the policy's readiness to react swiftly to changing conditions. 

It's worthwhile to note that this type of reactive behavior is acquired by the agent through interacting with the environment. In other words, the agent develops this behavior ``for free" - we don't explicitly formulate the problem to solve for a specific dynamic behavior. Instead, the desired response emerges naturally as part of the solution through ongoing interactions.
Previous work~\citet{marcucci2017hybrid, hogan2016feedback, aceituno-cabezas2020rss} have attempted to formulate these contact-rich manipulation problems as mixed-integer programming for the resulting hybrid systems, which allows the policy to plan different modes of contact and the accommodating motions. However, these methods can quickly become computationally intractable as the planning horizon increases, since the number of possible contact modes grows exponentially with the length of the planning horizon.
Additionally, they require accurate state estimators, which are not always available for many real-world tasks.

In contrast, our method directly leverages perception to learn reactive behaviors upon encountering contact. Through interaction, it encodes essential dynamics required to find a solution, rather than treating these dynamics as part of the problem formulation. Prior approaches, however, incorporate complex or intractable dynamics within the problem formulation itself, making these solutions harder to derive and less scalable.

Overall, our approach demonstrates the flexibility to learn these distinct policy types within a unified algorithmic framework. By interacting with the environment and observing the outcomes of their actions, the method adapts to the specific demands of each task. This adaptability enables the system to effectively address tasks that require diverse behaviors, spanning a wide range of manipulation challenges.

\section{Discussion}\label{sec:discuss}
The presented results substantially advance the published state-of-the-art in robotic manipulation. 
Our research demonstrates that with the right design choices, model-free RL can actually effectively tackle a variety of complex manipulation tasks using perception inputs, directly training in the real world within a practical timeframe. Trained policies from this approach are highly performant, achieving nearly perfect success rates that substantially exceed those of alternative approaches, such as imitation learning, along with cycle times that are also considerably faster.

Beyond the results themselves, the approach presented in this work can have significant broader impact. It can serve as a general framework for acquiring a wide range of manipulation skills with high performance and adapt to variations. This is particularly valuable in High-Mix Low-Volume (HMLV) manufacturing, or ``make-to-order" production~\cite{Jina1997ApplyingLP,shah2003129,app13031687}. Such production methods have substantial potential in major industries such as electronics, semiconductors, automotive, and aerospace due to their need for shorter product life cycles, customization, agility, and flexibility. 

We see a number of opportunities for future work. First, our approach can serve as an effective tool for generating high-quality data to train robot foundation models~\citep{rt1,rt2,rtx,octo,openvla}. Given that each task requires a relatively short time to train and the training process is largely autonomous, this framework can be employed to develop a variety of skills. Subsequently, data can be collected by executing the converged policies, which can then be distilled into these generalist models.
Second, although the current training time is relatively short, each task still requires training from scratch. We can further reduce this time by pretraining a value function that encapsulates the general manipulation capabilities of solving a range of different tasks with distinct robot embodiments. This pretrained value function can then be quickly fine-tuned to address specific tasks. 

We also see some limitations of our approach. Although we successfully address a variety of challenging tasks, it remains uncertain whether this method can be further extended to tasks with significantly longer horizons, where the sample complexity issue becomes more pronounced. However, this challenge might be alleviated through improved pretraining techniques or by employing methods that automatically segment a long-horizon task into a series of shorter sub-tasks, such as a vision-language model. 
It's also important to note that we did not perform extensive randomization in our experiments, nor did we test the method's generalization capability in unstructured environments. The primary focus of this paper is to demonstrate that the approach can be general-purpose in acquiring a wide range of manipulation skills with high performance. We believe that the randomization issue could be addressed by extending the training duration of the policies with the desired randomization level as in~\citet{Luo-RSS-21}. Additionally, the generalization issue might be resolved by incorporating vision foundation models that are pretrained on large-scale diverse datasets.

We hope this work will pave the way for the use of reinforcement learning in solving robotic manipulation problems, achieving high performance and eventually deploying them into the real world.

\clearpage

\section*{Acknowledgments}
The authors would like to thank Kyle Stachowicz and Qiyang Li for very helpful discussions.
\paragraph*{Funding:}
This research was partly supported by the AI Institute, ONR N00014-20-1-2383, and NSF IIS-2150826.
\paragraph*{Author contributions:}
J.L. designed the research, conceived the idea and devised the main method in the paper, implemented the prototype and performed initial experiments, provided hands-on guidance on all experiments, analyzed and interpreted the results, led the project, wrote and positioned the paper. 
C.X. contributed to the research design, maintained the main research codebase, prepared experiment hardware, performed a significant amount of experiments, prepared figures and videos for the paper.
J.W. contributed to the research design, maintained the main research codebase, performed a large amount of experiments, cleaned up the codebase for public release.
S.L. contributed to the research design, advised the project, edited and positioned the paper.
\paragraph*{Competing interests:}
There are no competing interests to declare.
\paragraph*{Data and materials availability:}
Accompanying videos and code can be found on the project website: \url{https://hil-serl.github.io/}. 

\newpage
\balance

\bibliography{main}

\newpage

\appendix

\section*{\Large \textbf{Supplementary Materials}}
\section{Task Setup and Policy Training Details}
In this section, we provide details regarding how each task is set up, including hardware and software; as well as details on policy training.

\subsection{RAM Insertion}
\begin{figure}[htbp]
    \centering
    \includegraphics[width=\textwidth]{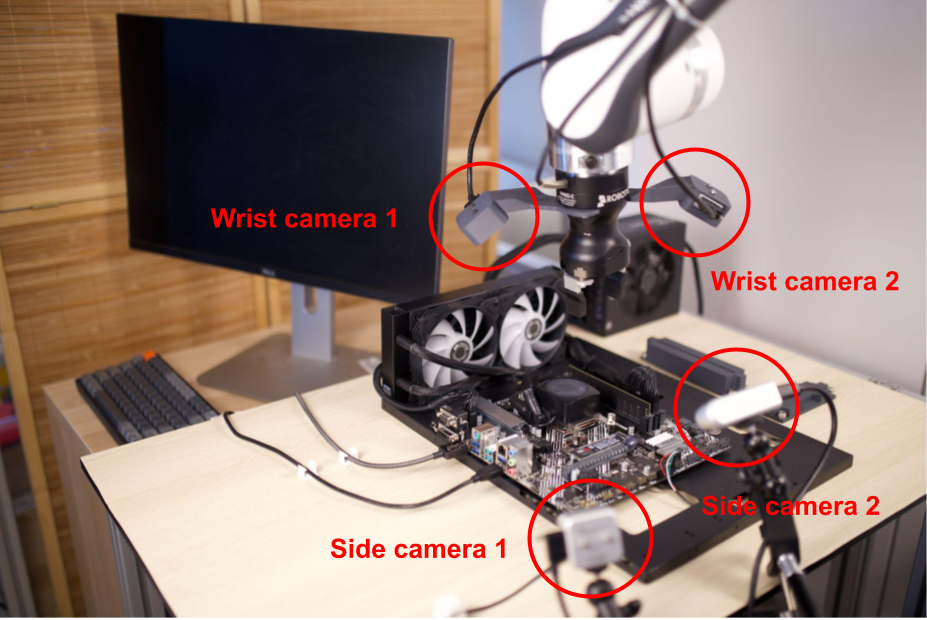}
    \caption{Hardware setup for the motherboard assembly task.} 
    \label{fig:sup_motherboard_setup}
\end{figure}

Fig.~\ref{fig:sup_motherboard_setup} shows the hardware setup for the motherboard assembly task, which presents the robot, the camera placements, and the task arrangement. 

\subsubsection{Cropped Images}
\begin{figure}[htbp]
    \centering
    \includegraphics[width=0.5\textwidth]{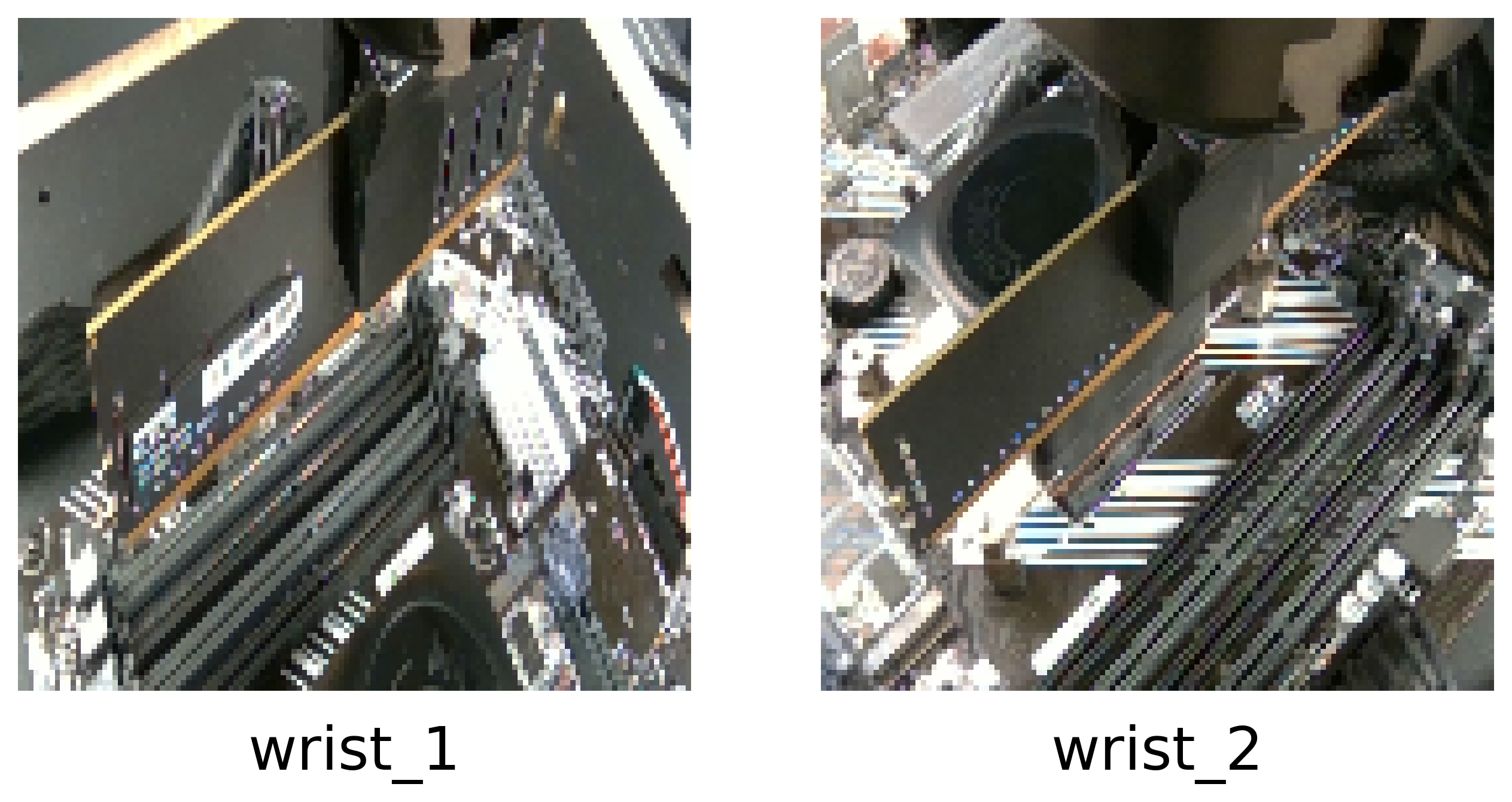}
    \caption{Sample input images from cameras used as inputs to the policy.
    } 
    \label{fig:sup_ram_cropped}
\end{figure}

We cropped the images to focus on the task-relevant parts of the scene, as shown in Fig.~\ref{fig:sup_ram_cropped}.

\subsubsection{Policy Training Details}
In Table~\ref{sup:ram}, we report additional details of the policy training for this task.

\begin{table}[H]
    \centering
    \begin{tabular}{l | l}
    %\hline
    Parameter & Value \\
    \hline
    Observation space & wrist\_1, wrist\_2, tcp\_pose, tcp\_vel, tcp\_f/t\\
    Action space & 6D twist  \\
    Reward function & Binary classifier \\
    Classifier views & wrist\_1, wrist\_2  \\
    Classifier accuracy & 97\% \\
    Initial offline demonstrations & 20\\
    Environment update frequency & 10 HZ \\
    Max episode length & 100 environment steps \\
    Reset method  & Scripted reset \\
    Randomization range & 4 cm in x and y, 6 deg in rz \\
    Proprio encoder size & 64 \\
    Policy MLP size & 256x256 \\
    Total number of RL transitions & 32000 \\
    Discount factor & 0.97 \\
    Optimizer & Adam \\
    Learning rate & 3e-4 \\
    Image augmentation & Random crop \\
    %\hline
    \end{tabular}
    \caption{Policy training details for the RAM insertion task.}
    \label{sup:ram}
\end{table}

\clearpage % This ensures all floats are processed before continuing

\subsection{SSD Assembly}
Fig.~\ref{fig:sup_motherboard_setup} shows the hardware setup for the motherboard assembly task, which presents the robot, the camera placements, and the task arrangement.

\subsubsection{Cropped Images}
\begin{figure}[htbp]
    \centering
    \includegraphics[width=\textwidth]{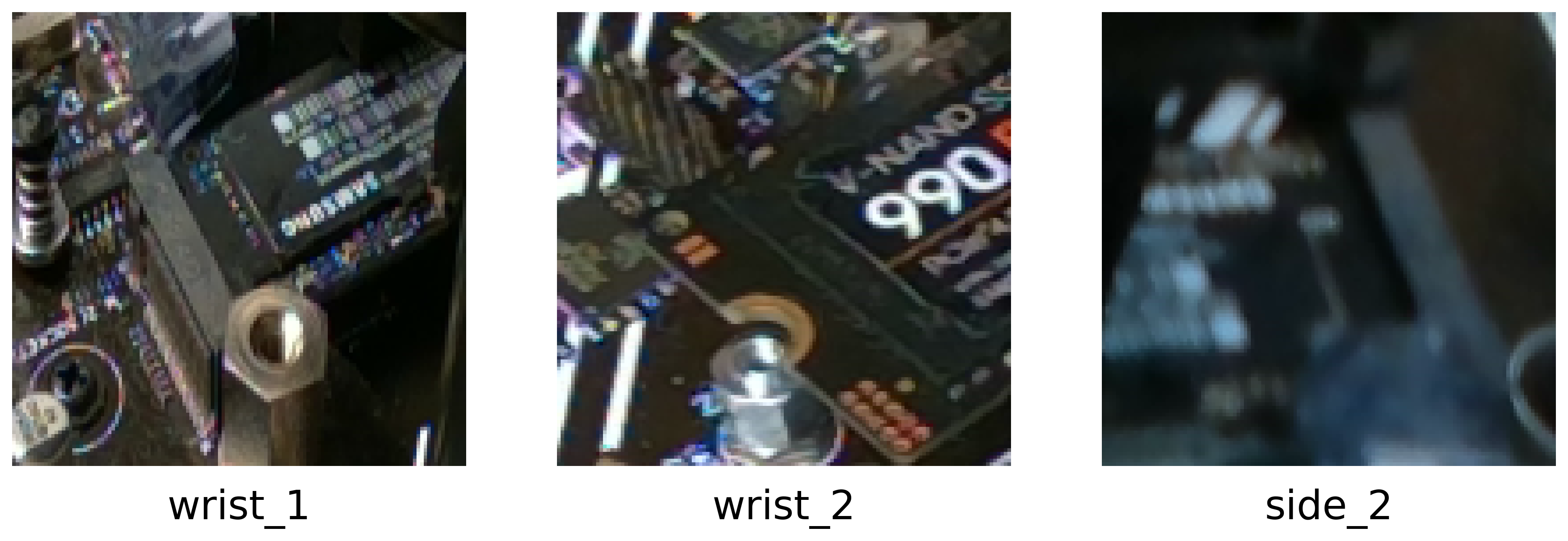}
    \caption{Sample input images from cameras used as inputs to the policy.
    } 
    \label{fig:sup_ssd_cropped}
\end{figure}

We cropped the images to focus on the task-relevant parts of the scene, as shown in Fig.~\ref{fig:sup_ssd_cropped}.

\subsubsection{Policy Training Details}
In Table~\ref{sup:ssd}, we report additional details of the policy training for this task.

\begin{table}[H]
    \centering
    \begin{tabular}{l | l}
    %\hline
    Parameter & Value \\
    \hline
    Observation space & wrist\_1, wrist\_2, side\_2, tcp\_pose, tcp\_vel, tcp\_f/t \\
    Action space & 6D twist \\
    Reward function & Binary classifier \\
    Classifier views & wrist\_1, wrist\_2, side\_2  \\
    Classifier accuracy & 95\% \\
    Initial offline demonstrations & 20\\
    Environment update frequency & 10 HZ \\
    Max episode length & 100 environment steps \\
    Reset method  & Scripted reset \\
    Randomization range & 2 cm in x and y, 1 deg in rz \\
    Proprio encoder size & 64 \\
    Policy MLP size & 256x256 \\
    Total number of RL transitions & 21000 \\
    Discount factor & 0.97 \\
    Optimizer & Adam \\
    Learning rate & 3e-4 \\
    Image augmentation & Random crop \\
    %\hline
    \end{tabular}
    \caption{Policy training details for the SSD assembly task.}
    \label{sup:ssd}
\end{table}

\clearpage % This ensures all floats are processed before continuing

\subsection{USB Grasp-Insertion}
Fig.~\ref{fig:sup_motherboard_setup} shows the hardware setup for the motherboard assembly task, which presents the robot, the camera placements, and the task arrangement. 

\subsubsection{Cropped Images}
\begin{figure}[htbp]
    \centering
    \includegraphics[width=\textwidth]{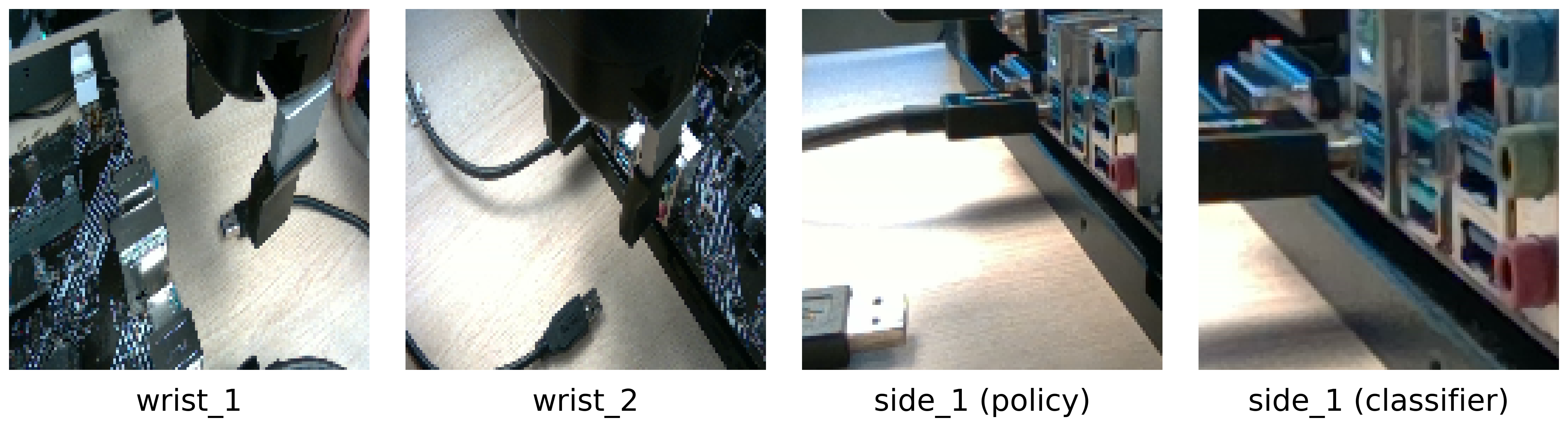}
    \caption{Sample input images from cameras used as inputs to the policy.
    } 
    \label{fig:sup_usb_insert_cropped}
\end{figure}

We cropped the images to focus on the task-relevant parts of the scene, as shown in Fig.~\ref{fig:sup_usb_insert_cropped}.

\subsubsection{Policy Training Details}
In Table~\ref{sup:usb}, we report additional details of the policy training for this task.

\begin{table}[H]
    \centering
    \begin{tabular}{l | l}
    %\hline
    Parameter & Value \\
    \hline
    Observation space & wrist\_1, wrist\_2, side\_1, tcp\_pose, tcp\_vel, tcp\_f/t, gripper\_pos \\
    Action space & 6D twist and 1D discrete gripper control \\
    Reward function & Binary classifier \\
    Classifier views & side\_1  \\
    Classifier accuracy & 96\% \\
    Initial offline demonstrations & 20\\
    Environment update frequency & 10 HZ \\
    Max episode length & 120 environment steps \\
    Reset method  & Scripted reset \\
    Randomization range & 2 cm in x and y, 10 deg in rz \\
    Proprio encoder size & 64 \\
    Motion policy MLP size & 256x256 \\
    Grasp critic MLP size & 256x256 \\
    Total number of RL transitions & 50000 \\
    Discount factor & 0.98 \\
    Optimizer & Adam \\
    Learning rate & 3e-4 \\
    Image augmentation & Random crop \\
    %\hline
    \end{tabular}
    \caption{Policy training details for the USB grasp and insertion task.}
    \label{sup:usb}
\end{table}

\clearpage % This ensures all floats are processed before continuing

\subsection{Cable Clipping}
Fig.~\ref{fig:sup_motherboard_setup} shows the hardware setup for the motherboard assembly task, which presents the robot, the camera placements, and the task arrangement.

\subsubsection{Cropped Images}
\begin{figure}[htbp]
    \centering
    \includegraphics[width=0.6\textwidth]{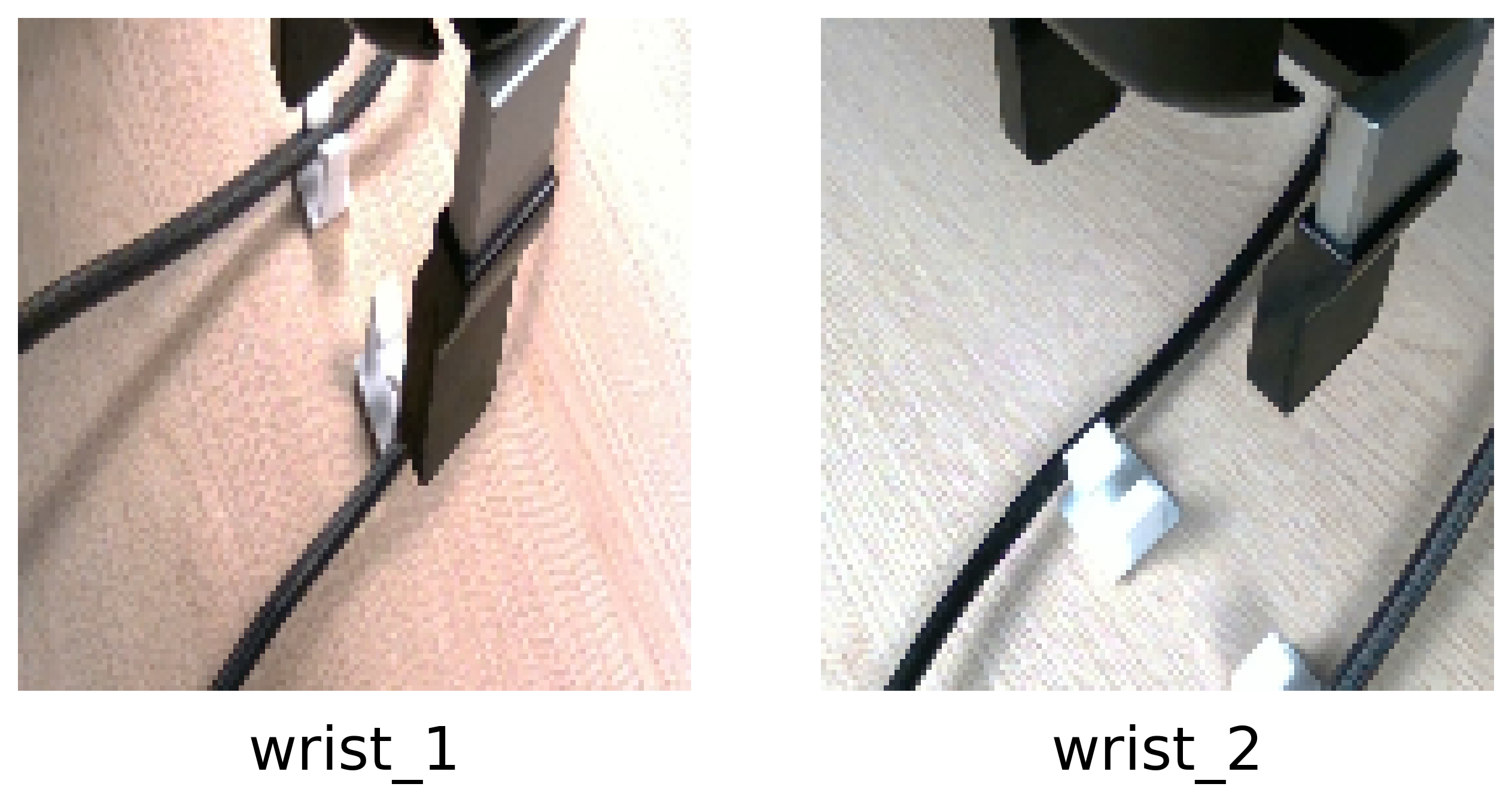}
    \caption{Sample input images from cameras used as inputs to the policy.
    } 
    \label{fig:sup_usb_route_cropped}
\end{figure}

We cropped the images to focus on the task-relevant parts of the scene, as shown in Fig.~\ref{fig:sup_usb_route_cropped}.

\subsubsection{Policy Training Details}
In Table~\ref{sup:cable}, we report additional details of the policy training for this task.

\begin{table}[H]
    \centering
    \begin{tabular}{l | l}
    %\hline
    Parameter & Value \\
    \hline
    Observation space & wrist\_1, wrist\_2,  tcp\_pose, tcp\_vel, tcp\_f/t, gripper\_pos \\
    Action space & 6D twist and 1D discrete gripper control \\
    Reward function & Binary classifier \\
    Classifier views & wrist\_1, wrist\_2  \\
    Classifier accuracy & 97\% \\
    Initial offline demonstrations & 20\\
    Environment update frequency & 10 HZ \\
    Max episode length & 120 environment steps \\
    Reset method  & Human reset \\
    Randomization range & 4 cm in x and y, 10 deg in rz \\
    Proprio encoder size & 64 \\
    Motion policy MLP size & 256x256 \\
    Grasp critic MLP size & 256x256 \\
    Total number of RL transitions & 28000 \\
    Discount factor & 0.98 \\
    Optimizer & Adam \\
    Learning rate & 3e-4 \\
    Image augmentation & Random crop \\
    %\hline
    \end{tabular}
    \caption{Policy training details for the cable clipping task.}
    \label{sup:cable}
\end{table}

\clearpage % This ensures all floats are processed before continuing

\subsection{IKEA - Side Panel}
\begin{figure}[htbp]
    \centering
    \includegraphics[width=\textwidth]{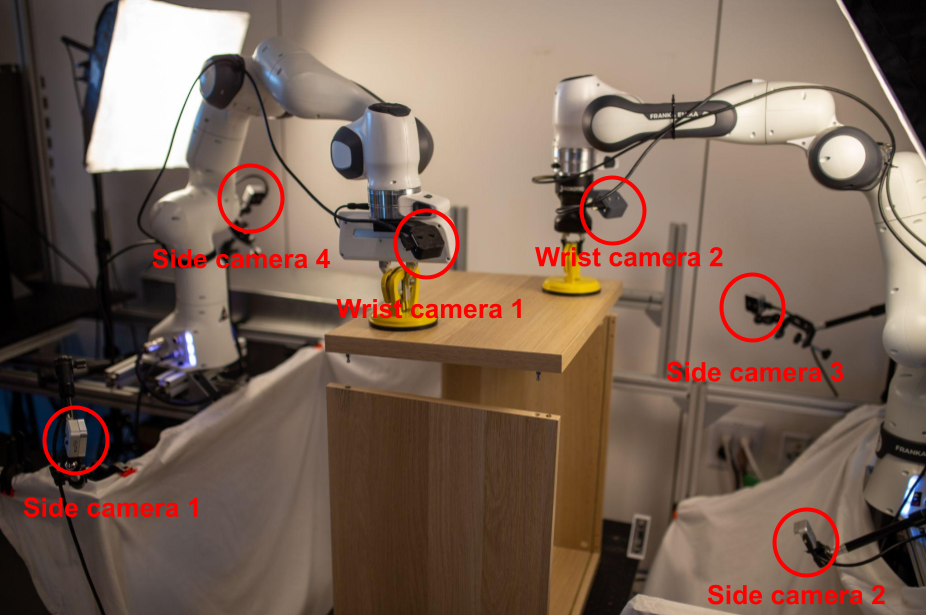}
    \caption{Hardware setup for the IKEA furniture assembly task.} 
    \label{fig:sup_ikea_setup}
\end{figure}

Fig.~\ref{fig:sup_ikea_setup} shows the hardware setup for the IKEA assembly task, which presents the robot, the camera placements, and the task arrangement. 

\subsubsection{Policy Training Details}
In Table~\ref{sup:ikea-side}, we report additional details of the policy training for this task.

\begin{table}[H]
    \centering
    \begin{tabular}{l | l}
    %\hline
    Parameter & Value \\
    \hline
    Observation space for side panel 1 & wrist\_1, side\_1, side\_2, tcp\_pose, tcp\_vel, tcp\_f/t \\
    Observation space for side panel 2 & wrist\_2, side\_3, side\_4, tcp\_pose, tcp\_vel, tcp\_f/t \\
    Action space & 12D twist \\
    Reward function & Binary Classifier \\
    Classifier views for panel 1 & side\_1,  side\_2 \\
    Classifier views for panel 2 & side\_3,  side\_4 \\
    Classifier accuracy & 97\% \\
    Initial offline demonstrations & 20\\
    Environment update frequency & 10 HZ \\
    Max episode length & 100 environment steps \\
    Reset method  & Scripted reset \\
    Randomization range & 8 cm in x, y, 1 deg in rz \\
    Proprio encoder size & 64 \\
    Policy MLP size & 256x256 \\
    Total number of RL transitions for panel 1 & 31000 \\
    Total number of RL transitions for panel 2 & 36000 \\
    Discount factor & 0.98 \\
    Optimizer & Adam \\
    Learning rate & 3e-4 \\
    Image augmentation & Random crop \\
    %\hline
    \end{tabular}
    \caption{Policy training details for the IKEA side panel task.}
    \label{sup:ikea-side}
\end{table}

\clearpage % This ensures all floats are processed before continuing

\subsection{IKEA - Top Panel}
Fig.~\ref{fig:sup_ikea_setup} shows the hardware setup for the IKEA assembly task, which presents the robot, the camera placements, and the task arrangement.

\subsubsection{Policy Training Details}
In Table~\ref{sup:ikea-top}, we report additional details of the policy training for this task.

\begin{table}[H]
    \centering
    \begin{tabular}{l | l}
    %\hline
    Parameter & Value \\
    \hline
    Observation space & side\_1, side\_3, side\_4, tcp\_pose, tcp\_vel, tcp\_f/t \\
    Action space & 12D twist \\
    Reward function & Binary Classifier \\
    Classifier views & side\_1, side\_3, side\_4  \\
    Classifier accuracy & 95\% \\
    Initial offline demonstrations & 20\\
    Environment update frequency & 10 HZ \\
    Max episode length & 150 environment steps \\
    Reset method  & Scripted reset \\
    Randomization range & 3 cm in x, y \\
    Proprio encoder size & 64 \\
    Policy MLP size & 256x256 \\
    Total number of RL transitions & 18000 \\
    Discount factor & 0.97 \\
    Optimizer & Adam \\
    Learning rate & 3e-4 \\
    Image augmentation & Random crop \\
    %\hline
    \end{tabular}
    \caption{Policy training details for the IKEA top panel task.}
    \label{sup:ikea-top}
\end{table}

\clearpage % This ensures all floats are processed before continuing

\subsection{Car Dashboard Assembly}
\begin{figure}[htbp]
    \centering
    \includegraphics[width=\textwidth]{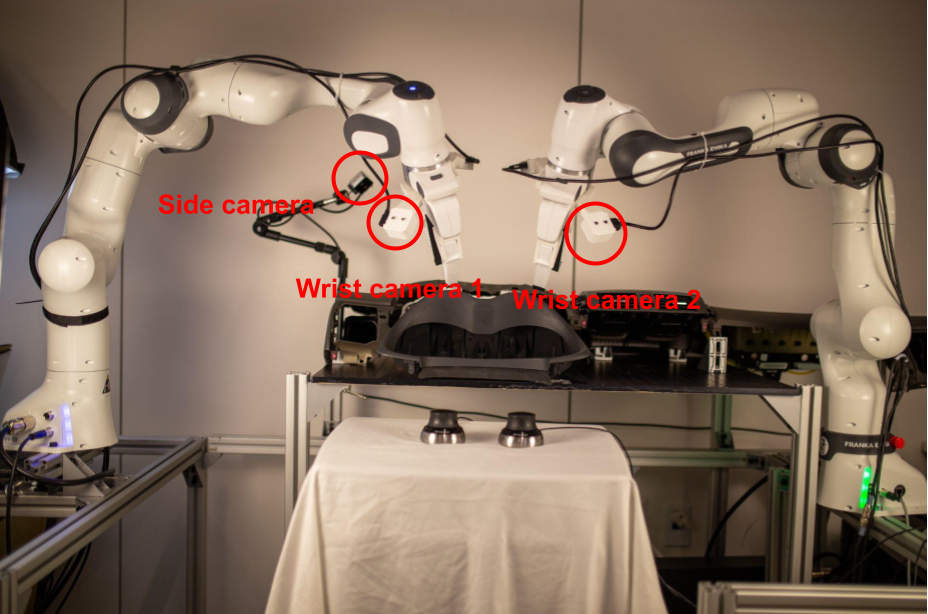}
    \caption{Hardware setup for the car dashboard installation task.} 
    \label{fig:sup_dashboard_setup}
\end{figure}

Fig.~\ref{fig:sup_dashboard_setup} shows the hardware setup for the dashboard installation task, which presents the robot, the camera placements, and the task arrangement. 

\subsubsection{Cropped Images}
\begin{figure}[htbp]
    \centering
    \includegraphics[width=\textwidth]{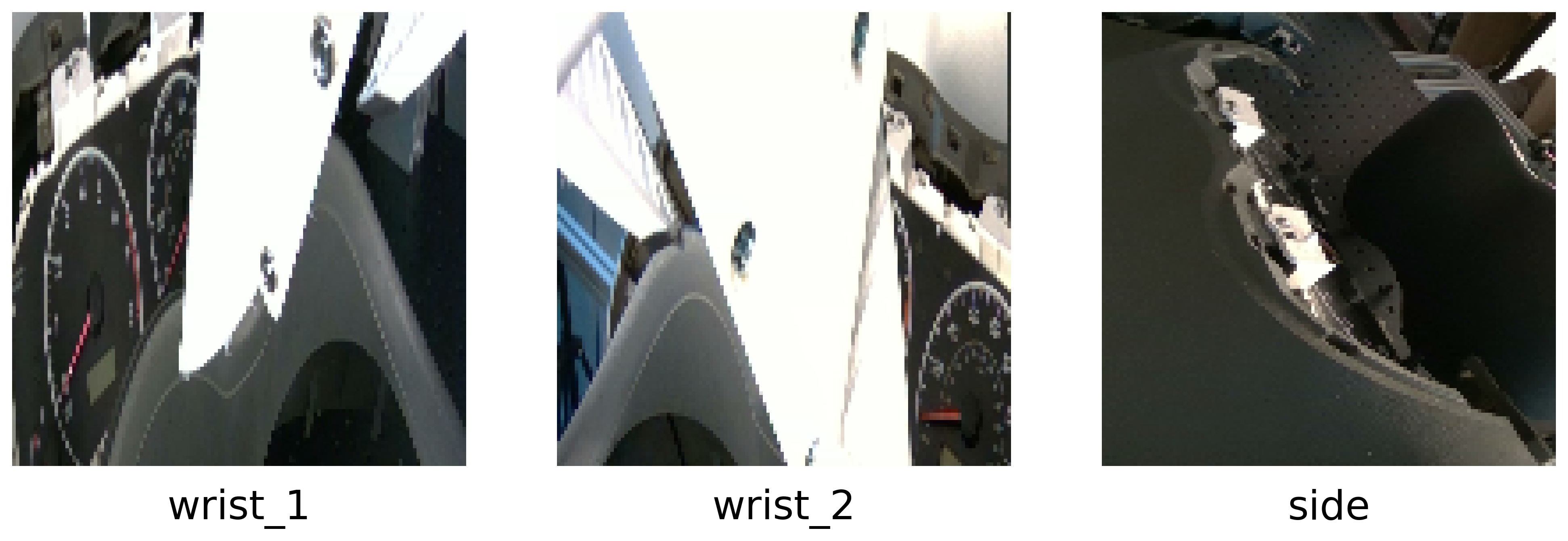}
    \caption{Sample input images from cameras used as inputs to the policy.
    } 
    \label{fig:sup_dash_cropped}
\end{figure}

We cropped the images to focus on the task-relevant parts of the scene, as shown in Fig.~\ref{fig:sup_dash_cropped}.

\subsubsection{Policy Training Details}
In Table~\ref{sup:dashboard}, we report additional details of the policy training for this task.

\begin{table}[H]
    \centering
    \begin{tabular}{l | l}
    %\hline
    Parameter & Value \\
    \hline
    Observation space & wrist\_1, wrist\_2, side, tcp\_pose, tcp\_vel, tcp\_f/t, gripper\_pos \\
    Action space & 12D twist and 1D discrete gripper control \\
    Reward function & Binary classifier \\
    Classifier views & wrist\_1, wrist\_2, side \\
    Classifier accuracy & 98\% \\
    Initial offline demonstrations & 20\\
    Environment update frequency & 10 HZ \\
    Max episode length & 200 environment steps \\
    Reset method  & Human reset \\
    Randomization range & 2 cm in x and y \\
    Proprio encoder size & 64 \\
    Motion policy MLP size & 256x256 \\
    Grasp critic MLP size & 256x256 \\
    Total number of RL transitions & 36000 \\
    Discount factor & 0.97 \\
    Optimizer & Adam \\
    Learning rate & 3e-4 \\
    Image augmentation & Random crop \\
    %\hline
    \end{tabular}
    \caption{Policy training details for the car dashboard assembly task.}
    \label{sup:dashboard}
\end{table}

\clearpage % This ensures all floats are processed before continuing

\subsection{Object Handover}
\begin{figure}[htbp]
    \centering
    \includegraphics[width=\textwidth]{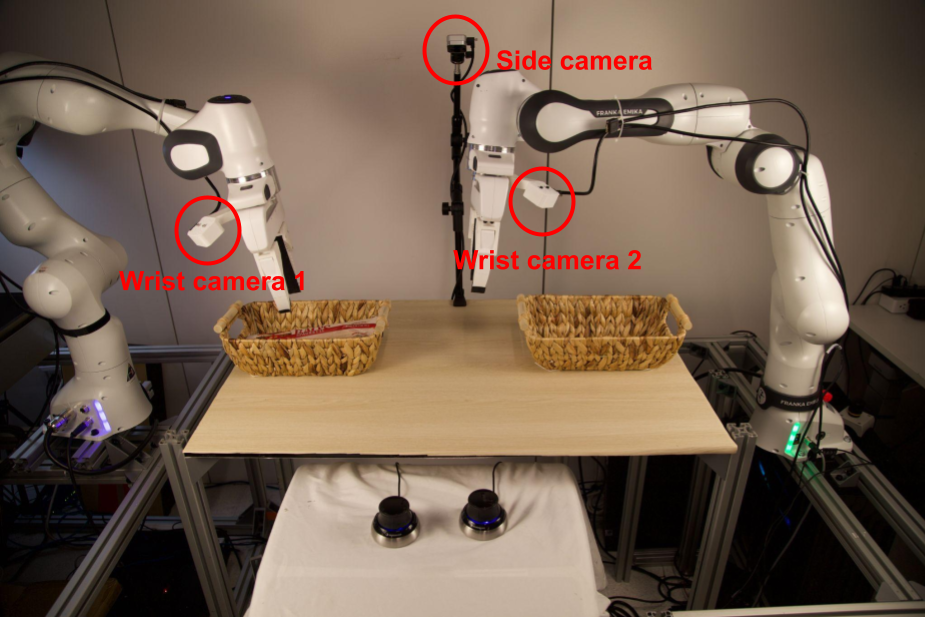}
    \caption{Hardware setup for the object handover task.} 
    \label{fig:sup_handover_setup}
\end{figure}

Fig.~\ref{fig:sup_handover_setup} shows the hardware setup for the object handover task, which presents the robot, the camera placements, and the task arrangement. 

\subsubsection{Cropped Images}
\begin{figure}[htbp]
    \centering
    \includegraphics[width=\textwidth]{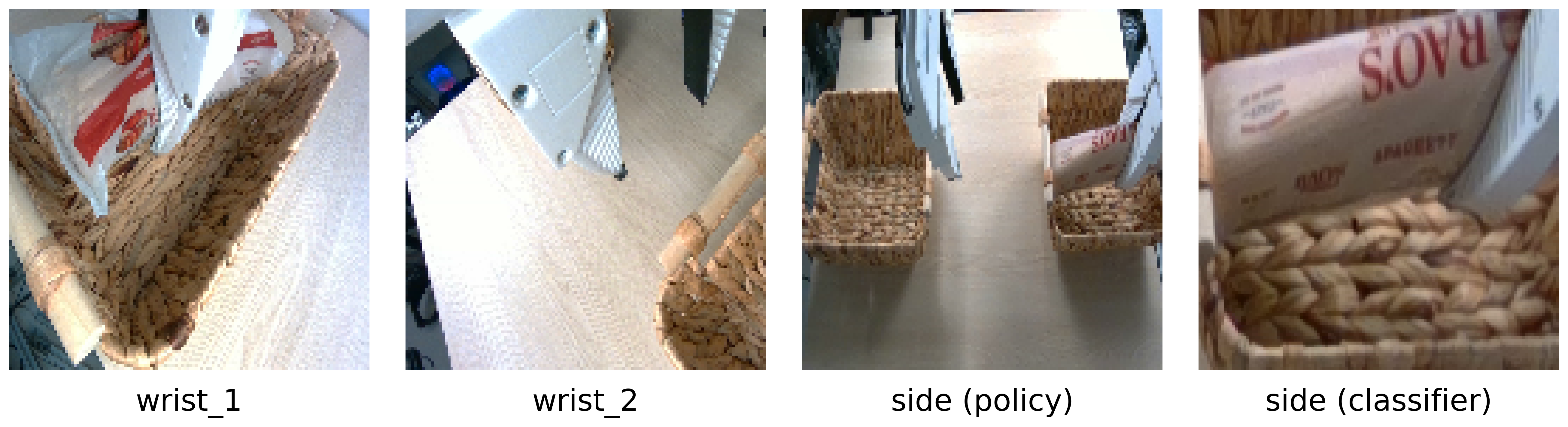}
    \caption{Sample input images from cameras used as inputs to the policy.
    } 
    \label{fig:sup_handover_cropped}
\end{figure}

We cropped the images to focus on the task-relevant parts of the scene, as shown in Fig.~\ref{fig:sup_handover_cropped}.

\subsubsection{Policy Training Details}
In Table~\ref{sup:object_handover}, we report additional details of the policy training for this task.

\begin{table}[H]
    \centering
    \begin{tabular}{l | l}
    %\hline
    Parameter & Value \\
    \hline
    Observation space & wrist\_1, wrist\_2, side, tcp\_pose, tcp\_vel, gripper\_pos \\
    Action space & 12D twist and 1D discrete gripper control\\
    Reward function & Binary classifier \\
    Classifier views & side  \\
    Classifier accuracy & 99\% \\
    Initial offline demonstrations & 20\\
    Environment update frequency & 10 HZ \\
    Max episode length & 200 environment steps \\
    Reset method  & Human reset \\
    Randomization range & None \\
    Proprio encoder size & 64 \\
    Motion policy MLP size & 256x256 \\
    Grasp critic MLP size & 256x256 \\
    Total number of RL transitions & 43000 \\
    Discount factor & 0.97 \\
    Optimizer & Adam \\
    Learning rate & 3e-4 \\
    Image augmentation & Random crop \\
    %\hline
    \end{tabular}
    \caption{Policy training details for the object handover task.}
    \label{sup:object_handover}
\end{table}

\clearpage % This ensures all floats are processed before continuing

\subsection{Timing Belt Assembly}
\begin{figure}[htbp]
    \centering
    \includegraphics[width=\textwidth]{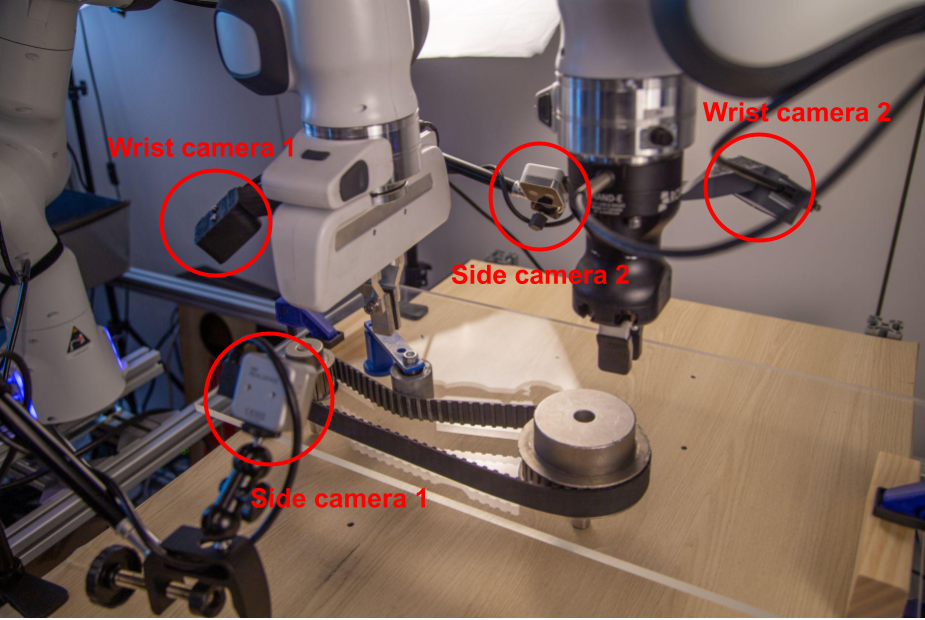}
    \caption{Hardware setup for the timing belt assembly task.} 
    \label{fig:sup_timing_belt_setup}
\end{figure}

Fig.~\ref{fig:sup_timing_belt_setup} shows the hardware setup for the timing belt assembly task, which presents the robot, the camera placements, and the task arrangement. 

\subsubsection{Cropped Images}
\begin{figure}[htbp]
    \centering
    \includegraphics[width=\textwidth]{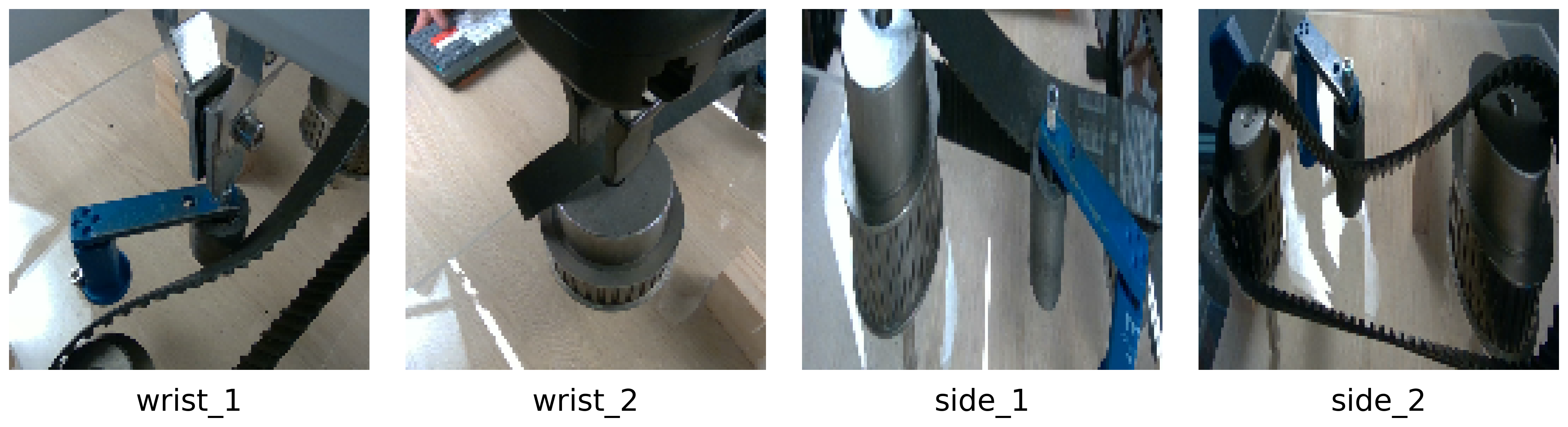}
    \caption{Sample input images from cameras used as inputs to the policy.
    } 
    \label{fig:sup_timing_belt_cropped}
\end{figure}

We cropped the images to focus on the task-relevant parts of the scene, as shown in Fig.~\ref{fig:sup_timing_belt_cropped}.

\subsubsection{Policy Training Details}
In Table~\ref{sup:timingbelt}, we report additional details of the policy training for this task.

\begin{table}[H]
    \centering
    \begin{tabular}{l | l}
    %\hline
    Parameter & Value \\
    \hline
    Observation space & wrist\_1, wrist\_2, side\_1, side\_2, tcp\_pose, tcp\_vel, tcp\_f/t \\
    Action space & 12D twist \\
    Reward function & Binary classifier \\
    Classifier views & side\_1, side\_2  \\
    Classifier accuracy & 96\% \\
    Initial offline demonstrations & 20\\
    Environment update frequency & 10 HZ \\
    Max episode length & 200 environment steps \\
    Reset method  & Human reset \\
    Randomization range & 2 cm in x and y \\
    Proprio encoder size & 64 \\
    Policy MLP size & 256x256 \\
    Total number of RL transitions & 108000 \\
    Discount factor & 0.97 \\
    Optimizer & Adam \\
    Learning rate & 3e-4 \\
    Image augmentation & Random crop \\
    %\hline
    \end{tabular}
    \caption{Policy training details for the timing belt assembly task.}
    \label{sup:timingbelt}
\end{table}

\clearpage % This ensures all floats are processed before continuing

\subsection{Jenga Whipping}
\begin{figure}[htbp]
    \centering
    \includegraphics[width=\textwidth]{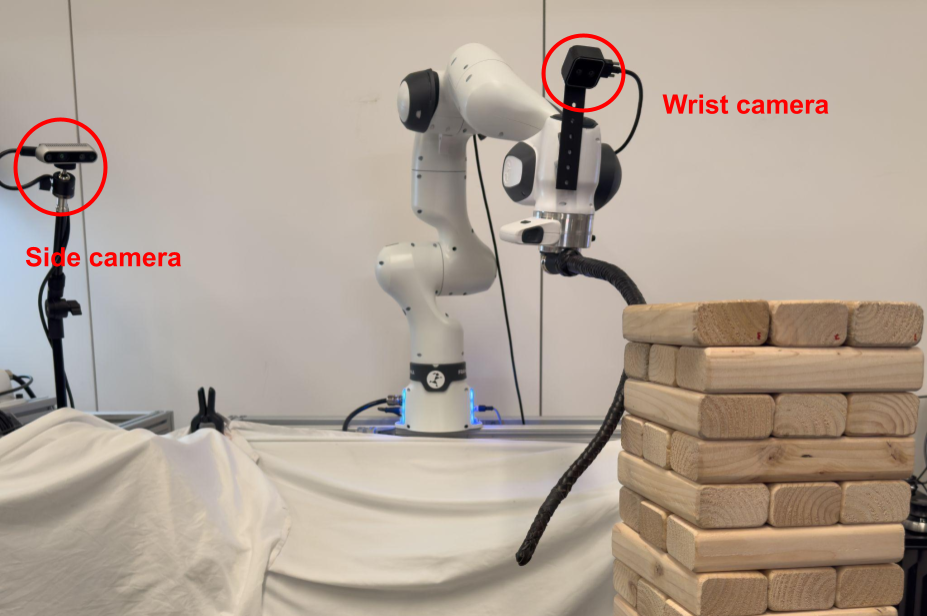}
    \caption{Hardware setup for the Jenga whipping task.} 
    \label{fig:sup_jenga_setup}
\end{figure}

Fig.~\ref{fig:sup_jenga_setup} shows the hardware setup for the Jenga whipping task, which presents the robot, the camera placements, and the task arrangement. 

\subsubsection{Cropped Images}
\begin{figure}[htbp]
    \centering
    \includegraphics[width=0.5\textwidth]{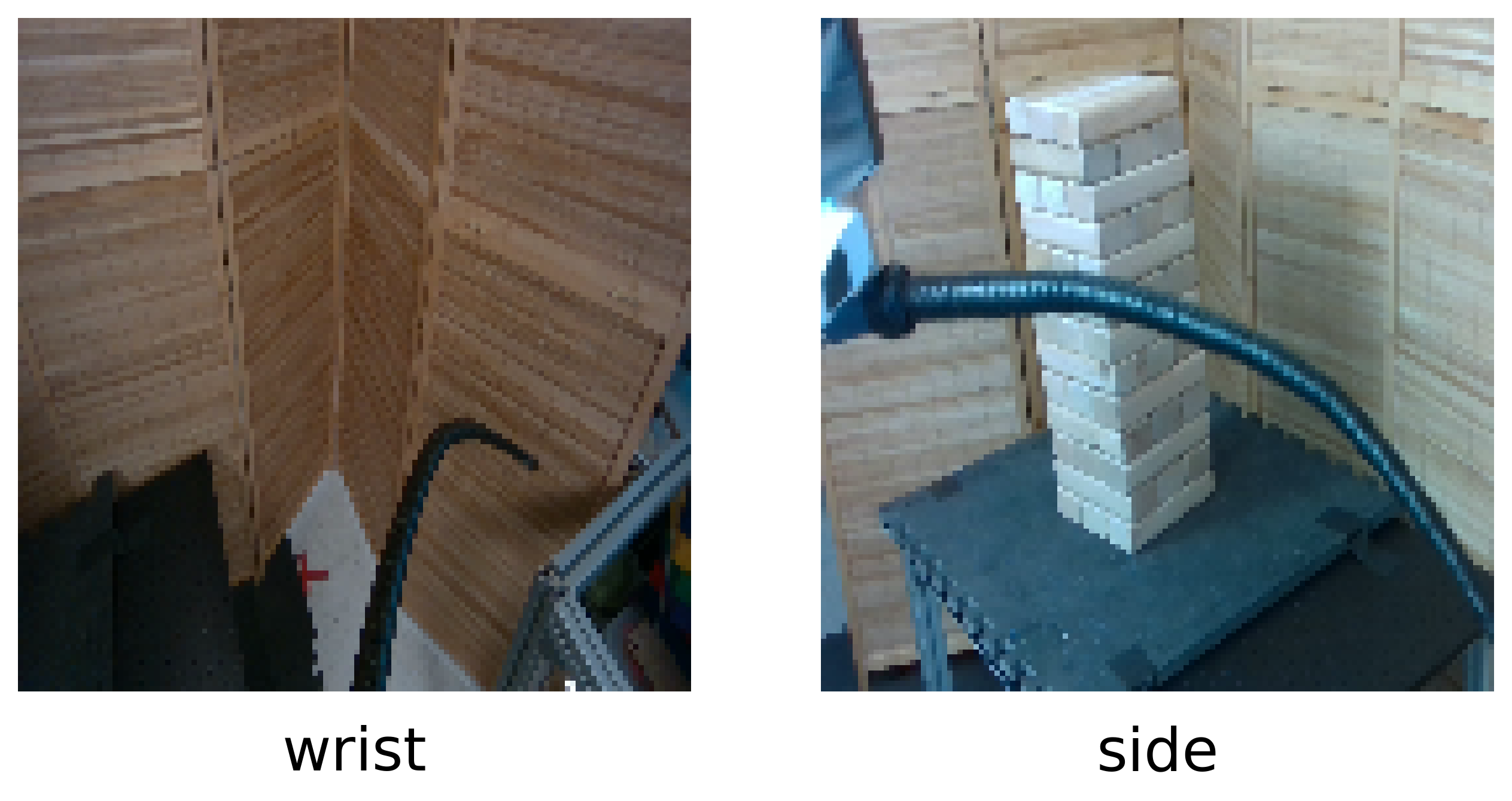}
    \caption{Sample input images from cameras used as inputs to the policy.
    } 
    \label{fig:sup_jenga_cropped}
\end{figure}

We cropped the images to focus on the task-relevant parts of the scene, as shown in Fig.~\ref{fig:sup_jenga_cropped}.

\subsubsection{Policy Training Details}
In Table~\ref{sup:jenga}, we report additional details of the policy training for this task.

\begin{table}[H]
    \centering
    \begin{tabular}{l | l}
    %\hline
    Parameter & Value \\
    \hline
    Observation space & wrist, side, tcp\_pose, tcp\_vel, q, dq \\
    Action space & Feedforward wrench $F_x$, $F_z$, $\tau_z$ \\
    Reward function & Human annotation in the end of an episode \\
    Environment update frequency & 10 HZ \\
    Max episode length & 20 environment steps \\
    Reset method  & Human reset \\
    Randomization range & None \\
    Initial offline demonstrations & 30\\
    Proprio encoder size & 64 \\
    Policy MLP size & 256x256 \\
    Total number of RL transitions & 10000 \\
    Discount factor & 0.96, but every episode was run to maximum length \\
    Optimizer & Adam \\
    Learning rate & 3e-4, decayed to 3e-5 when reaching 70\% success rate \\
    Image augmentation & Random crop \\
    %\hline
    \end{tabular}
    \caption{Policy training details for the Jenga whipping task.}
    \label{sup:jenga}
\end{table}

\clearpage % This ensures all floats are processed before continuing

\subsection{Object Flipping}
\begin{figure}[htbp]
    \centering
    \includegraphics[width=0.5\textwidth]{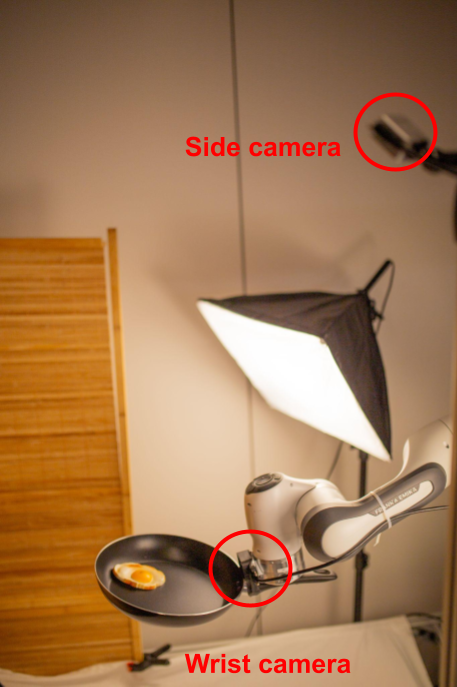}
    \caption{Hardware setup for the object flipping task.} 
    \label{fig:sup_object_flipping_setup}
\end{figure}

Fig.~\ref{fig:sup_object_flipping_setup} shows the hardware setup for the object flipping task, which presents the robot, the camera placements, and the task arrangement. 

\subsubsection{Cropped Images}
\begin{figure}[H]
    \centering
    \includegraphics[width=0.5\textwidth]{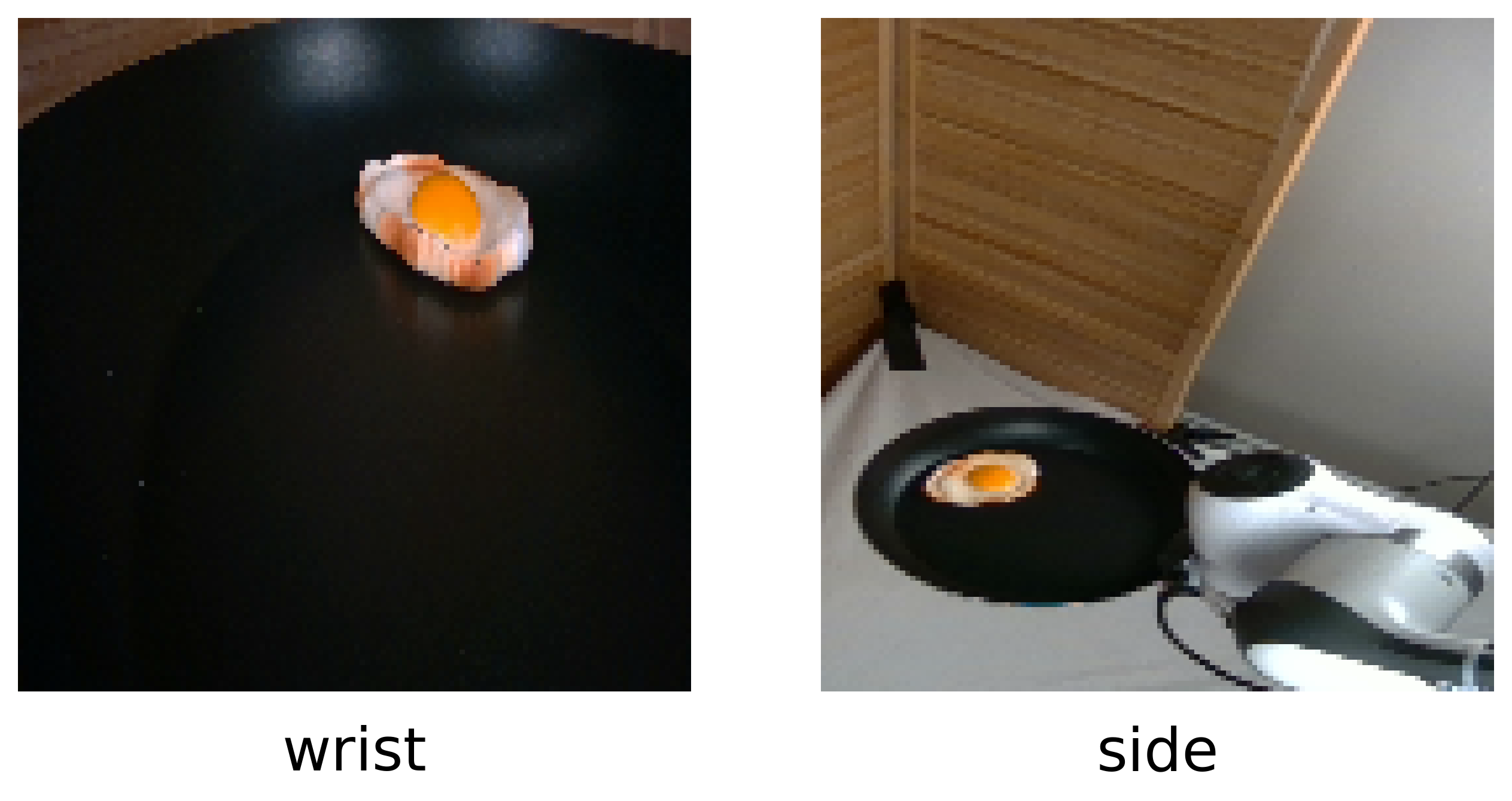}
    \caption{Sample input images from cameras used as inputs to the policy.
    } 
    \label{fig:sup_egg_flip_cropped}
\end{figure}

We cropped the images to focus on the task-relevant parts of the scene, as shown in Fig.~\ref{fig:sup_egg_flip_cropped}.

\subsubsection{Policy Training Details}
In Table~\ref{sup:flipping}, we report additional details of the policy training for this task.

\begin{table}[H]
    \centering
    \begin{tabular}{l | l}
    %\hline
    Parameter & Value \\
    \hline
    Observation space & wrist, side, tcp\_pose, tcp\_vel, q, dq \\
    Action space & Feedforward wrench $F_x$, $F_z$, $\tau_y$  \\
    Reward function & Binary classifier \\
    Classifier views & wrist  \\
    Classifier accuracy & 97\% \\
    Initial offline demonstrations & 20\\
    Environment update frequency & 10 HZ \\
    Max episode length & 100 environment steps \\
    Reset method  & Scripted reset \\
    Randomization range & None \\
    Proprio encoder size & 64 \\
    Policy MLP size & 256x256 \\
    Total number of RL transitions & 25000 \\
    Discount factor & 0.985 \\
    Optimizer & Adam \\
    Learning rate & 3e-4 \\
    Image augmentation & Random crop \\
    %\hline
    \end{tabular}
    \caption{Policy training details for the object flipping task.}
    \label{sup:flipping}
\end{table}

\clearpage % This ensures all floats are processed before continuing

\section{Reward Classifier Training Details}
For the reward classifiers, we used a pre-trained ResNet-10 model as the feature extractor, and connected it to a two-layer MLP, and we then trained the network on the collected dataset with cross-entropy loss. The classifier was trained using the Adam optimizer with a learning rate of 3e-4. The total number of training iterations was 100.

\begin{figure}[H]
    \centering
    \includegraphics[width=0.6\linewidth]{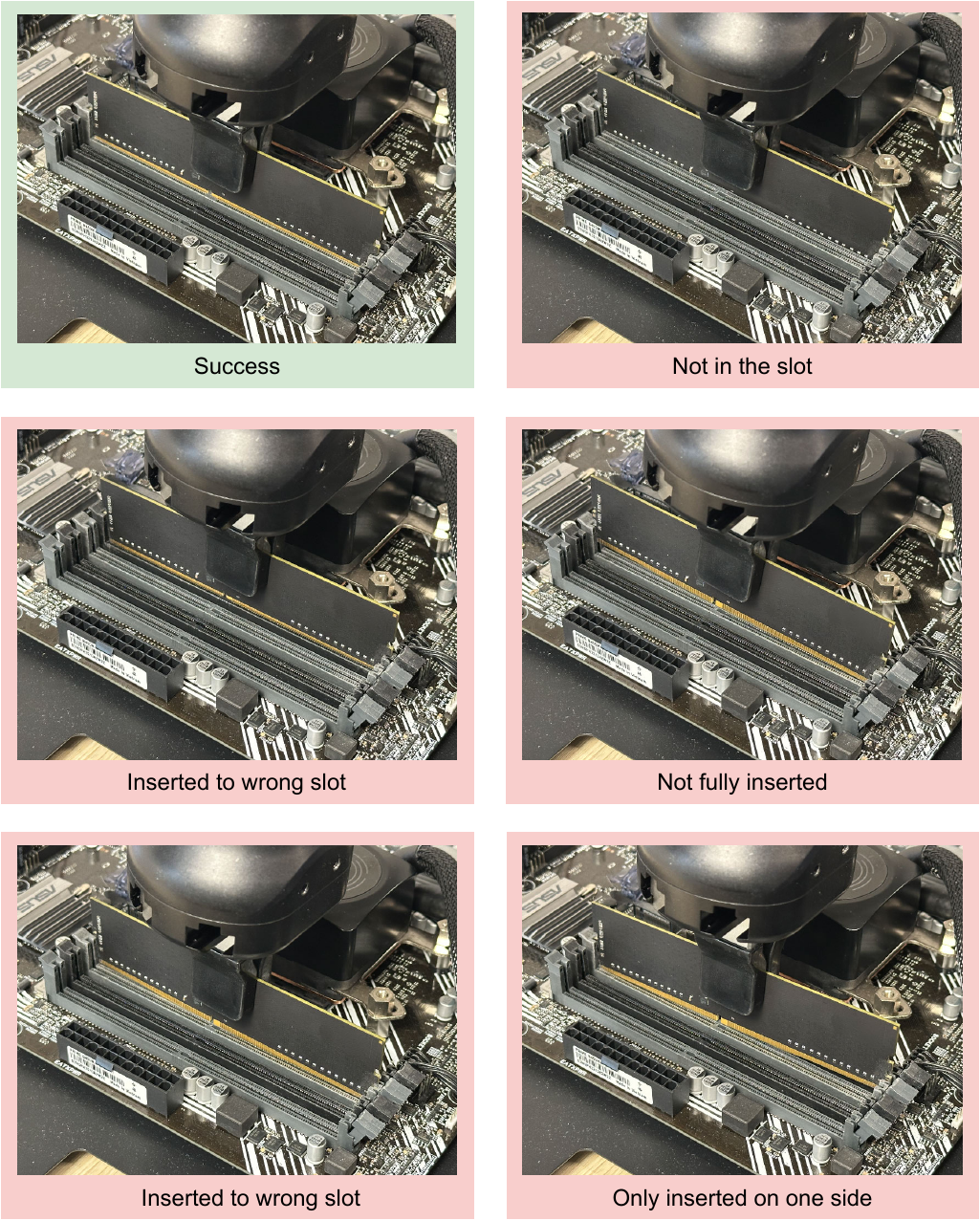}
    \caption{Sample images collected to train the reward classifier for the RAM insertion task.}
    \label{fig:reward_classifier_examples}
\end{figure}

To collect the training dataset, we teleoperated the robot to perform the task and recorded the images and labels with a SpaceMouse. We clicked the SpaceMouse button when the robot successfully completed the task, and marked those images with labels as 1. Otherwise, we marked the labels as 0. In some of the tasks, we also recorded additional false positive and false negative samples to improve the classifier performance. We represent a few such examples in Fig.~\ref{fig:reward_classifier_examples} to help readers understand how to train such classifiers.

\clearpage

\section{Diffusion Policy Training Details}
In this section, we provide detailed parameters for how we train the diffusion policy baselines, presented in Table.~\ref{sup:diffusion}.

\begin{table}[H]
    \centering
    %\begin{tabular}{ l l l l l}
    
    \begin{tabular}{ c c c c c }
    \hline    
    \multicolumn{1}{c}{$\begin{substack}{\text{Task}\\\text{Name}}\end{substack}$} & \multicolumn{1}{c}{$\begin{substack}{\text{Number of}\\\text{Demos}}\end{substack}$} & \multicolumn{1}{c}{$\begin{substack}{\text{Observation}\\\text{Chunking Size}}\end{substack}$} & \multicolumn{1}{c}{$\begin{substack}{\text{Action Prediction}\\\text{Horizon}}\end{substack}$} & \multicolumn{1}{c}{$\begin{substack}{\text{Action}\\\text{Chunking Size}}\end{substack}$} \\
    \hline
    \small{RAM Insertion} & 200 & 1 & 8 & 2 \\
    
    \small{Dashboard Assembly} & 200 & 1 & 8&4 \\
    
    \small{Object Flipping} & 200 & 1 & 1&1 \\
    
    \hline
    \end{tabular}
    \caption{Diffusion policy training details.}
    \label{sup:diffusion}
\end{table}

\clearpage

\section{Robot Controller and Proprioceptive Information Representation}
In this section, we detail the implementation of the robot controller and the representation of the proprioceptive information for the robots.

\subsection{Proprioceptive Information Representation}
Let the robot's base frame be $\{s\}$; for the $i$-th episode of rolling out the policy, we denote $\{b_t^{(i)}\}$ as the end-effector frame expressed w.r.t. $\{s\}$ at a particular time step $t$; where $1\leq i \leq M$, $0\leq t \leq N$. For each episode, $\{b_0^{(i)}\}$ is sampled from a uniform distribution specifying the area of randomization. We want to express such proprioceptive information with respect to $\{b_0^{(i)}\}$. Thus, the policy will be applicable to a new location provided that the relative spatial distance between the robot's end-effector and the target remains consistent. This approach prevents overfitting to specific global locations within the reference frame $\{s\}$.
We achieve this by applying the following homogeneous transformation:
$$T_{b_0^{(i)}b_t^{(i)}} = T_{b_0^{(i)}}^{-1} \cdot T_{b_t^{(i)}}$$
where we use $T_{ab}$ to denote the homogeneous transformation matrix between frame $\{a\}$ and $\{b\}$. We feed the position and rotation information extracted from $T_{b_0^{(i)}b_t^{(i)}}$ to the policy.
Here we use $T_{ab}$ to denote the homogeneous transformation matrix between frame $\{a\}$ and $\{b\}$, defined as:

\begin{equation}
  T_{ab} =   \begin{bmatrix}
R_{ab} & p_{ab} \\
0_{1 \times 3} & 1 
\end{bmatrix}.
\end{equation}

For most of the tasks, the policy generates a six-degree-of-freedom (6 DoF) twist action, which is expressed in the reference frame from which it currently receives observations, i.e., $\{b_t^{(i)}\}$. 
Mathematically, the 6 DoF twist action $\mathcal{V}_t^{(i)}$ is expressed in frame $\{b_t^{(i)}\}$ at timestep $t$. 
To interface with the robot's control software, which expects actions ${\mathcal{V}_t^{(i)}}^{\prime}$ expressed in the base frame $\{s\}$, we apply the adjoint mapping:
$${\mathcal{V}_t^{(i)}}^{\prime} = {[\text{Ad}_t^{(i)}]}\mathcal{V}_t^{(i)}$$
where $[\text{Ad}_t^{(i)}]$ is a function of the homogeneous transformation $T_{b_t^{(i)}}$ defined as:
\begin{equation}
  [\text{Ad}_t^{(i)}] =   \begin{bmatrix}
R_{b_t^{(i)}} & 0_{3 \times 3} \\
[p_{b_t^{(i)}}]\times R_{b_t^{(i)}} & R_{b_t^{(i)}}
\end{bmatrix}.
\end{equation}

For two dynamic manipulation tasks, the policy generates a 3 DoF feedforward wrench action, which is also expressed in the reference frame from which it currently receives observations, i.e., $\{b_t^{(i)}\}$. It will then be sent to the low-level robot controller as setpoints, which then will be converted to joint torques for execution by multiplying the transpose of the Jacobian matrix at the current timestep $t$.

\subsection{Robot Controller}
For most of the tasks, the low-level robot controller is an impedance controller running at 1000 Hz, which accepts 10 Hz setpoints computed by the policy. As discussed in~\cite{luo2024serl}, we perform additional treatment on it to ensure the stability of the training process in most contact-rich manipulation tasks. Consider a typical impedance controller without feedforward term: 
\begin{equation}
    F = k_p \cdot e + k_d \cdot \dot{e} + F_{ff} + F_{cor},
\end{equation}
where $e = p - p_{ref}$, $p$ is the measured pose, and $p_{ref}$ is the target pose computed by the upstream controller, $F_{ff}$ is the desired feedforward force, $F_{cor}$ is the Coriolis force. This objective will then be converted into joint space torques by multiplying Jacobian transpose and offset by nullspace torques. It acts like a spring-damper system around the equilibrium set by $p_{ref}$ with the stiffness coefficient being $k_p$ and the damping coefficient being $k_d$. As described above, this system will yield large forces if $p_{ref}$ is far away from the current pose, which can lead to a hard collision or damage when the arm is in contact with something. Therefore it's crucial to constrain the interaction force generated by it. However, directly reducing gains will hurt the controller's accuracy. Thus, we should bound $e$ so that $|e| \leq \Delta$,  and then the generated force from the spring-damper system will be bounded to $k_p \cdot |\Delta | + 2 k_d \cdot |\Delta| \cdot f$, $f$ is the control frequency. 

For the two dynamic manipulation tasks, we run a feedforward wrench controller at 1000 Hz, which accepts 10 Hz setpoints computed by the policy. It converts the desired wrench into joint torques by multiplying the Jacobian transpose and offset by nullspace torques.

\clearpage

\section{Policy Training Plots}
In this section, we provide additional plots for HIL-SERL policy training for all tasks.

\begin{figure}[H]
    \centering
    \includegraphics[width=1\linewidth]{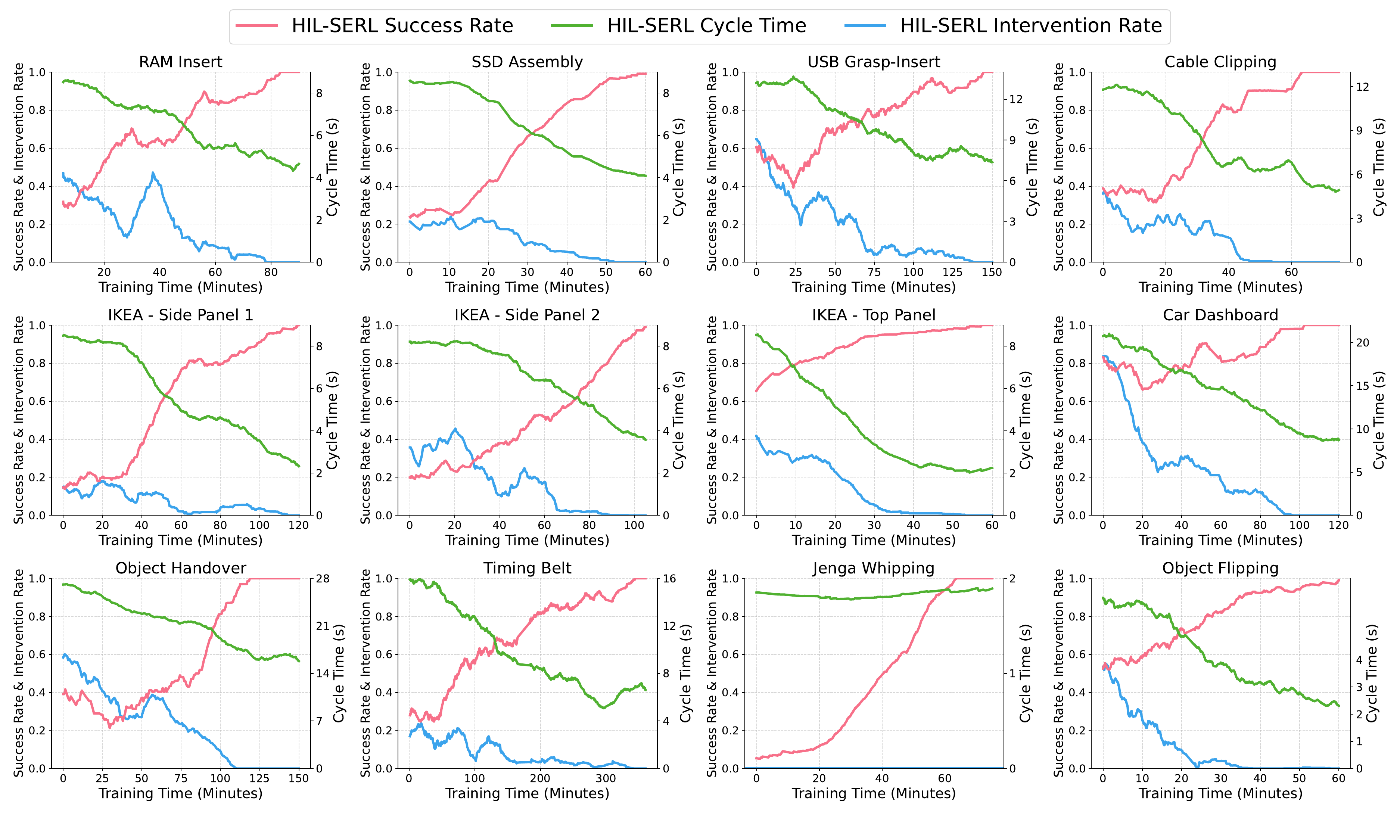}
    \caption{\textbf{Learning curves for experimental tasks.} This figure presents the success rate, cycle time, and intervention rates for both HIL-SERL across all experiment tasks, displayed as a running average over 20 episodes. 
    The success rate increased rapidly throughout training, eventually reaching 100\%, while the intervention rate and cycle time progressively decreased, with the intervention rate ultimately reaching 0\%.}
    \label{sup:training_plot}
\end{figure}

\end{document}